\documentclass{article} %
\usepackage[preprint]{colm2026_conference} %

\usepackage{microtype}
\usepackage{hyperref}
\usepackage{url}
\usepackage{booktabs}

\usepackage{amsmath,amsfonts,bm}

\def\eqref#1{equation~\ref{#1}}

\def\1{\bm{1}}

\DeclareMathAlphabet{\mathsfit}{\encodingdefault}{\sfdefault}{m}{sl}
\SetMathAlphabet{\mathsfit}{bold}{\encodingdefault}{\sfdefault}{bx}{n}

\newcommand{\R}{\mathbb{R}}

\usepackage{annotation}

\usepackage{graphicx}
\usepackage{wrapfig}
\usepackage{tikz}
\usepackage{xcolor}
\usepackage{amssymb}
\usepackage[nameinlink]{cleveref}
\usepackage{xpatch}
\usepackage{enumitem}
\usepackage{picinpar}
\usepackage{xcolor}

\usepackage{ai-usage-card}

\aiProjectName{Mind the Gap Between Spatial Reasoning and Acting! Step-by-Step Evaluation of Agents With \sparcgym}
\aiDomain{Paper}
\aiKeyApplication{Spatial Reasoning}

\aiContactName{Lars Benedikt Kaesberg}
\aiContactEmail{l.kaesberg@uni-goettingen.de}
\aiContactAffiliation{University Göttingen}

\aiModels{ChatGPT 5.2, Claude Opus 4.6, Gemini 3 Pro}

\aiFindingLiterature{Claude, ChatGPT}

\aiImprovingContent{Claude, Gemini}

\aiGeneratingCode{Claude}
\aiRefactoringCode{Claude}

\usepackage{lineno}

\definecolor{darkblue}{rgb}{0, 0, 0.5}
\hypersetup{colorlinks=true, citecolor=darkblue, linkcolor=darkblue, urlcolor=darkblue}

\title{Mind the Gap Between Spatial Reasoning and Acting!\\Step-by-Step Evaluation of Agents With \sparcgym{}}

\author{%
Lars Benedikt Kaesberg\quad
Tianyu Yang\quad
Niklas Bauer\quad
Terry Ruas\quad
Jan Philip Wahle\thanks{Equal contribution. Author order randomized by coin flip.}\quad
\\\bf
Bela Gipp\footnotemark[1]\smallskip\\
University of Göttingen, Germany \\
\texttt{l.kaesberg@uni-goettingen.de}
}

\newcommand{\sparcgym}{Spatial-Gym}

\begin{document}

\ifcolmsubmission
\linenumbers
\fi

\maketitle
\AddAnnotationRef{}
\vspace{-20pt}
\begin{abstract}
Spatial reasoning is central to navigation and robotics, yet measuring model capabilities on these tasks remains difficult.
Existing benchmarks evaluate models in a one-shot setting, requiring full solution generation in a single response, unlike humans, who work in interactive environments step-by-step.
We introduce \sparcgym{}, a Gymnasium environment that isolates spatial constraint reasoning by testing pathfinding in 2D-grid puzzles as a sequential decision task with optional backtracking.
We evaluate eight models in three settings (one-shot, step-by-step, step-by-step with backtracking) against human, random, and A* baselines on 500 episodes.
The best model, GPT-OSS 120B, achieves a solve rate of 16.0\%, 82 points below the human baseline (98.0\%).
Step-by-step format helps weaker models (up to +5.4\%) by removing formatting errors, but hurts stronger models (up to $-$5.6\%) by constraining global planning.
Backtracking improves episode completion, but increases solve rate only for weaker models; stronger models rarely backtrack and do not benefit from it.
Our experiments have three key findings: (1) models fail to scale reasoning effort with difficulty, (2) vision models receiving images of the spatial environment reduce solve rate by 73\%, and (3) extended chain-of-thought reasoning retains a $3-5\times$ accuracy advantage over standard inference even in the step-by-step setting.
\sparcgym{} enables diagnosis of model limitations and provides a framework for improving spatial reasoning through reinforcement learning.
\vspace{-10pt}

\end{abstract}

\section{Introduction}
\label{sec:intro}
\vspace{-5pt}
Spatial reasoning is a central human ability for navigating the real world and for designing some of humanity's most complex inventions, including robotic motion planners and printed circuit boards \citep{metastudyspatial}. 
Large language models (LLMs) are increasingly deployed as high-level planners for such tasks, generating decision sequences that maintain overarching goals \citep{ahn2022icanisay, song2023llmplannerfewshotgroundedplanning, Zhou_2025}. 
Yet it remains unclear how well these models can perform spatial reasoning, because current evaluations rely on one-shot, non-interactive tasks rather than step-by-step environments that capture how humans solve such problems. 
Most benchmarks require models to produce a complete solution in a single pass, making it difficult to assess where reasoning mistakes occur and to distinguish whether a failure reflects flawed reasoning or an inability to express a valid output \citep{aghzal2025largelanguagemodelsgood, dao2025alphamazeenhancinglargelanguage}.
A recently presented dataset for testing spatial reasoning \citep{kaesberg2025sparcspatialpathfindingreasoning} combines path planning with interacting constraints and shows a large human–model gap (98\% vs.\ 15.8\%), unaffected by architecture, scale, or prompting. 
Prior work on textual gridworlds \citep{kim2024languagemodelsextrapolateoutside} and maze navigation \citep{POUDEL2026100447} shows that when models perform stepwise reasoning, they improve performance, but these environments use simple rule constraints that may overlap with pretraining data, and do not support backtracking prior steps. 
\citet{qin2025backtrackbacktracksequentialsearch} show that backtracking can improve model performance on structured tasks such as Sudoku, but harms them in less constrained tasks such as Countdown \citep{katz2025seeminglysimpleplanningproblems}. 
Existing spatial reasoning tasks lack a combination of sequential decision-making, rule-based constraints, and backtracking in an interactive reinforcement learning (RL) environment.

\begin{figure}[t]
    \centering
    \includegraphics[width=\textwidth]{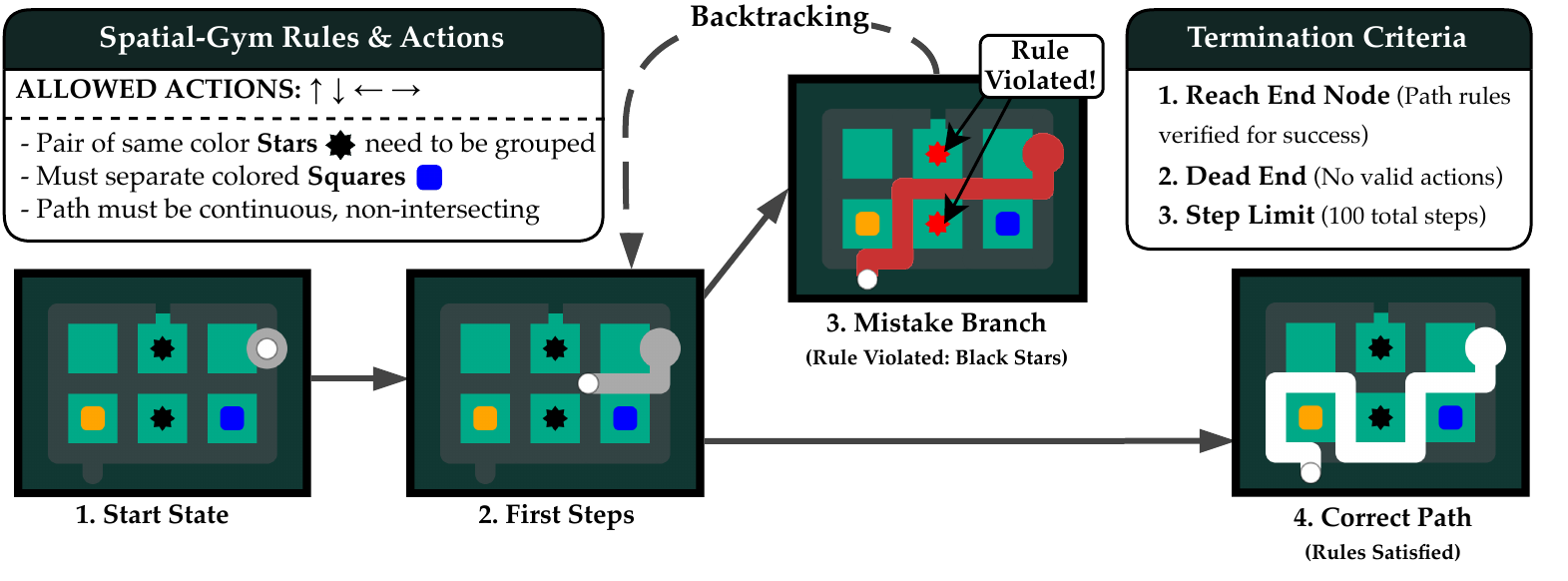}
    \vspace{-30pt}
    \caption{Overview of the \sparcgym{} task.}
    \label{fig:fig1}
\end{figure}

We introduce \textbf{\sparcgym{}} (cf. \Cref{fig:fig1}), a Gymnasium environment \citep{towers2025gymnasiumstandardinterfacereinforcement} that formulates 2D grid pathfinding with interacting constraints as a sequential decision task with automatic solution verification and outcome and process rewards for RL. 
The environment builds on \citet{kaesberg2025sparcspatialpathfindingreasoning} and provides 1{,}000 tasks across five difficulty levels. 
Agents select to construct one path tile at a time, given the current grid state, path history, and legal moves, with optional backtracking to undo prior steps. 

We evaluate eight models across three settings: one-shot, stepwise (\sparcgym{}), and stepwise with backtracking (\sparcgym{}+backtracking).
We compare models with human, random, and A* baselines (\S\ref{sec:experiments}). 
Our results reveal four key findings. 
First, the best model, GPT-OSS 120B, achieves 16.0\% accuracy in the Gym setting, 82 points below humans (98.0\%).
\sparcgym{} isolates spatial reasoning as the primary bottleneck in current models and shows that this limitation is independent of scale, architecture, and evaluation format. 
Second, we observe a trade-off between instruction-following of models to format the correct response and actual reasoning capabilities. 
The interactive environment format helps smaller models by removing formatting errors (up to +5.4\% accuracy) but hurts frontier models (up to~$-$5.6\% accuracy), possibly because their long reasoning traces can lead to long-context issues (\hyperref[sec:main_results]{\S\ref*{sec:main_results}}).
Third, backtracking reduces deadlocks (i.e., when no legal actions are available anymore) for all models but improves accuracy only for weaker models; stronger models rarely use it and lose performance when they do (\hyperref[sec:backtracking]{\S\ref*{sec:backtracking}}). This behavior suggests that model training incentives discourage exploration in constraint-heavy tasks. 
Fourth, models fail to scale reasoning effort with puzzle difficulty and use backtracking to shorten paths rather than explore alternatives.
Visual language models receiving the entire puzzle as an image perform generally worse than regular LLMs (\hyperref[sec:vision]{\S\ref*{sec:vision}}).

\paragraph{Key Contributions:}
\begin{itemize}[leftmargin=*,label={\color{teal}$\blacktriangleright$},itemsep=0.15em]

\item (\hyperref[sec:gym]{\S\ref*{sec:gym}}) \sparcgym{}, a Gymnasium environment that formulates 2D grid pathfinding with interacting rule constraints as a sequential decision task with optional backtracking. We also provide verification and reward signals for future RL methods. 

\item (\hyperref[sec:main_results]{\S\ref*{sec:main_results}}) An evaluation of eight models in one-shot, step-by-step, and step-by-step with backtracking against human, random, and A* baselines showing a 82-percentage-point human--model accuracy gap.

\item (\hyperref[sec:backtracking]{\S\ref*{sec:backtracking}}) The identification of a new formatting--reasoning tradeoff and a backtracking paradox in which the step-by-step format helps smaller models (up to +5.4\% accuracy) but hinders frontier ones (up to $-$5.6\% accuracy) while backtracking improves path completion but increases accuracy only for smaller models.

\item (\hyperref[sec:path_effort]{\S\labelcref{sec:path_effort,sec:rule_analysis,sec:ablation_baseline,sec:vision,sec:scaling}}) Additional experiments on path generation effort, rule type difficulty, algorithmic baselines, and spatial reasoning with visual language models.

\item A public release of code and data for evaluating frontier models and training better ones for spatial reasoning.\footnote{Code and data are anonymously available under: \href{https://github.com/lkaesberg/Spatial-Gym}{\sparcgym{}}
 and \href{https://github.com/lkaesberg/spatial-gym-analysis}{\sparcgym{}-Analysis}}

\end{itemize}

\section{Related Work}
\label{sec:related-work}

\textbf{Spatial reasoning.}
Early spatial datasets such as bAbI \citep{weston2015aicompletequestionansweringset}, SpartQA \citep{mirzaee2021spartqatextualquestion}, and StepGame \citep{shi2022stepgamenewbenchmarkrobust} are based on multiple-choice formats, commonsense knowledge, or highly predictable sentence structures, and not on complex spatial logic. 
Recent work raised the difficulty by introducing more complex abstract visual reasoning tasks (e.g., VisualPuzzles \citep{song2025visualpuzzlesdecouplingmultimodalreasoning}, ARC \citep{chollet2019measureintelligence, chollet2025arcprize2024technical}). 
\citet{aghzal2025largelanguagemodelsgood} proposed a benchmark on spatial-temporal reasoning and showed that LLM performance on grid-based path planning collapsed with grid size and obstacle count. 
The SPaRC dataset \citep{kaesberg2025sparcspatialpathfindingreasoning} combined path planning with interacting rule constraints and found a human--model performance gap that neither architecture, scale, representation, nor prompting could explain. 
However, all of these evaluations were one-shot (in the evaluation sense, not in the model-prompting sense). The model produced a single complete answer, which was scored against a (set of) ground truth solutions. 
This is not how humans typically solve these problems, and could conflate output formatting ability with spatial reasoning ability, motivating Gym-based environments.

\textbf{Stepwise spatial navigation and LLMs as agents.}
Reformulating spatial tasks as sequential decisions can mitigate the aforementioned issues and mimic how humans would solve such tasks \citep{Zamprogno2023AHC}. Highlighting this shift, the recently introduced ARC-AGI-3 \citep{foundation2026arcagi3newchallengefrontier} benchmark transitioned abstract reasoning into an interactive, turn-based environment in which agents must explore and efficiently plan action sequences. Similarly, \citet{kim2024languagemodelsextrapolateoutside} showed that cognitive-map-style chain-of-thought (CoT) \citep{wei2023chainofthoughtpromptingelicitsreasoning} (where individual steps were abstracted in the verbalization) improved extrapolation to larger grids in textualized gridworld tasks. \citet{dao2025alphamazeenhancinglargelanguage} combined supervised fine-tuning with GRPO \citep{shao2024deepseekmathpushinglimitsmathematical} for step-by-step maze solving, reaching 93\% accuracy. Both results showed that stepwise interaction helped, but the mazes were comparably simple, containing only walls and open cells without interacting rule constraints. Whether these gains transferred to constraint-heavy tasks remained an open question. Beyond spatial domains, language models were evaluated as sequential decision agents in web browsing \citep{zhou2024webarenarealisticwebenvironment}, code execution \citep{liu2025agentbenchevaluatingllmsagents}, and other interactive settings \citep{chen2025efficientsequentialdecisionmaking}.
\citet{qin2025backtrackbacktracksequentialsearch} found that backtracking improved performance on structured constraint tasks, but limited models on less constrained problems.
These studies motivate the need for an environment that operates in a step-by-step, interactive manner and supports the evaluation of more complex mechanisms, such as backtracking.

\textbf{This work.}
Recent Gymnasium-based platforms such as KORGym \citep{shi2025korgymdynamicgameplatform} and Reasoning Gym \citep{stojanovski2025reasoninggymreasoningenvironments} included spatial tasks as subcategories of broader reasoning benchmarks, but none focused on spatial constraint reasoning or provided interacting rule constraints and backtracking.
To the best of our knowledge, \sparcgym{} is the first Gymnasium environment dedicated to spatial constraint reasoning.
It reformulates 2D grid puzzles as a sequential decision process and eliminates path-construction errors by providing only valid moves at each state. %

\section{\sparcgym{}}
\label{sec:gym}

Humans do not generate complete solutions to puzzles in a single pass \citep{Zamprogno2023AHC}.
They make assessments, reason about the next step, and backtrack when needed.
Yet most constraint benchmarks evaluate models by requiring a complete solution in a single response, from logic grid puzzles \citep{lin2025zebralogicscalinglimitsllms} to combinatorial tasks \citep{mittal2025fcorebenchlargelanguagemodels,gevers2026hintbenchmarkingllmsboard}.

We propose \sparcgym{} to reframe spatial reasoning as sequential single-step decisions, matching how humans naturally solve such puzzles and isolating reasoning from formatting.
The Gym builds on \citet{kaesberg2025sparcspatialpathfindingreasoning} and consists of 1,000 2D grid pathfinding puzzles (500 train, 500 test) across five difficulty levels. 
Unlike \citet{kaesberg2025sparcspatialpathfindingreasoning}, it formulates the task as a Markov Decision Process via the Gymnasium interface, introducing step-by-step rule evaluation, partial rewards, and backtracking.
Each puzzle is an $m \times n$ grid of rule cells with a start and end point on the grid edges.
The goal is to draw a continuous, non-self-intersecting path along grid edges from start to end while satisfying all rule cell constraints. 
The Gym has seven rule types: \textit{Dots} (path must pass through), \textit{Gaps} (path cannot cross), \textit{Stones} (color separation), \textit{Stars} (pairing with same-color elements), \textit{Triangles} (edge counting), \textit{Polyominoes} (shape fitting), and \textit{Ylops} (shape subtraction). 
Puzzle difficulty is a weighted combination of distinct rule count, total rule cells, rule density, grid size, and estimated rule interactions, normalized to a 1–5 scale. 
Full rule definitions are available in \Cref{app:sparc_dataset}. 
We use the same 500 test puzzles from \citet{kaesberg2025sparcspatialpathfindingreasoning} for evaluation.

We formalize \sparcgym{} as a Markov Decision Process $(S,A,T,R)$ using the Gymnasium interface \citep{towers2025gymnasiumstandardinterfacereinforcement}.
Each state $s_t \in S$ consists of the current environment grid (including rule cells and their types), the path traced so far (as a sequence of visited edge nodes), the agent's current position, and the set of legal actions available from that position.
The action space is $A = \{\text{up}, \text{down}, \text{left}, \text{right}\}$. 
At each step, only actions leading to valid, unvisited adjacent nodes are offered to the agent.
This eliminates invalid-path errors, such as disconnected lines, self-intersections, rule-cell crossings, and incorrect start/end conditions. %
The transition function $T(s_{t+1}\mid s_t, a_t)$ defines how the environment updates the agent's position and extends the traced path.
The agent receives the updated board state as a text representation, including the grid with the previously walked path, a list of legal actions, and the current position coordinates.
An episode terminates when either of the following conditions is met: (1) the agent reaches the end node, at which point the path is evaluated against all rule conditions to determine success or failure; (2) the agent reaches a deadlock with no valid moves remaining; or (3) hits a 100-step limit. This prevents infinite loops without bottlenecking performance; only 3.9\% of episodes reach it, mostly during unproductive backtracking loops.
By default, the Gym setting prevents the agent from backtracking, but the Gym can enable a mode that allows previous actions to be reverted.

The reward function $R(s_t,a_t)$ provides two reward signals designed for RL training. 
The \textit{outcome reward} is $+1$ if the agent reaches the end node and all rule conditions are satisfied, and $-1$ otherwise (deadlock or step limit). 
The \textit{process reward} is $+0.01$ for each step that results in a path from start to the current location, which matches with a sub-path of the solution, and $-0.01$ for any other case.

\section{Experiments}
\label{sec:experiments}

We evaluate eight models: GPT-OSS 120B, OLMo 3.1 32B, Nemotron 49B, Qwen 3 32B, R1 Distill Qwen 32B, Gemma 3 27B, Magistral Small 24B, and Qwen 3 0.6B.
We test these models under three settings. 
Baseline (one-shot path generation), \sparcgym{} (step-by-step), and \sparcgym{} with backtracking.
We report the percentage of puzzles solved (accuracy) and the percentage of puzzles where the agent reached the end node (completion rate).
Details on model versions and hardware are provided in \Cref{app:models_hardware}.

\subsection{General Results}
\label{sec:main_results}

\begin{wrapfigure}[14]{r}{0.5\linewidth}
\vspace{-10pt}
\centering
\includegraphics[width=.96\linewidth]{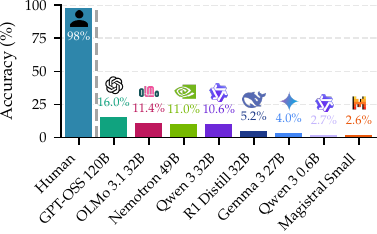}
\vspace{-7pt}
\caption{Accuracy (\%) of all models on \sparcgym{} with no backtracking.}
\label{fig:overall_accuracy}
\vspace{-15pt}
\end{wrapfigure}

We run all eight models on \sparcgym{} without backtracking on the full 500-puzzle test set.
Each model receives the system prompt with rules, the current board state, and legal actions at each step (see \Cref{app:prompts} for the full prompts and \Cref{app:example_gym} for a complete example with GPT-OSS 120B).
We plot the human baseline from \citet{kaesberg2025sparcspatialpathfindingreasoning}, in which humans also solved puzzles step by step via an interactive interface.

\Cref{fig:overall_accuracy} shows the accuracy values per model.
GPT-OSS 120B is the leading frontier model (16.0\%), followed by OLMo 3.1 32B (11.4\%), Nemotron 49B (11.0\%), and Qwen 3 32B (10.6\%).
Other models are below 6\%.

Two findings stand out.
First, three reasoning-trained 32B models (OLMo 3.1, Qwen 3, Nemotron) cluster between 10.6\% and 11.4\%, while R1 Distill 32B reaches only 5.2\%.
Unlike the other three, R1 Distill is trained through distillation from a larger teacher rather than direct RL on reasoning tasks \citep{deepseekai2025deepseekr1incentivizingreasoningcapability}, suggesting that the way models learn to reason matters more than parameter count within a size class.
Second, Gemma 3 27B, the only model without reasoning-specific tuning, outperforms the smaller reasoning models and almost matches R1 Distill 32B.
Full tabular results, including exact breakdowns by difficulty, rule type, and average step counts, are in \Cref{app:full_tabular_results}.
We also analyzed inter-model agreement and found that while frontier models solve a superset of puzzles, weaker models occasionally find unique solutions, suggesting multi-agent ensembles could be beneficial (details in \Cref{app:model_agreement,app:unique_solves}).

\begin{figure}[t]
\includegraphics[width=\linewidth]{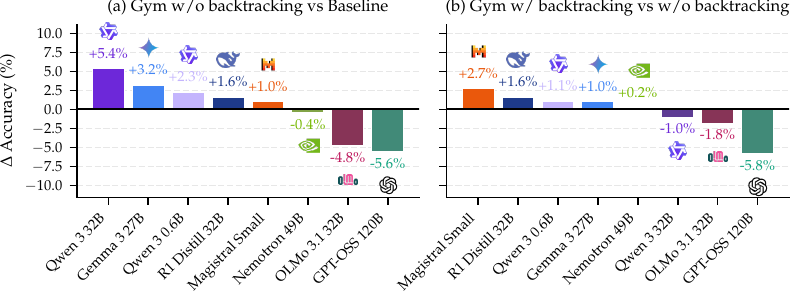}
\vspace{-18pt}
\caption{$\Delta$ Accuracy (\%) for each model relative to two settings: (a) Gym accuracy minus baseline accuracy; (b) Gym w/ backtracking accuracy minus w/o backtracking accuracy. Bars above zero show improvement under new conditions; bars below zero indicate collapse.}
\label{fig:delta}
\vspace{-15pt}
\end{figure}

\textbf{Formatting and reasoning tradeoff.} We compare each model's accuracy on the baseline task vs. \sparcgym{} on identical puzzles.
\Cref{fig:delta}a shows the accuracy delta of models (\sparcgym{} $-$ baseline). 
Delta values correlate inversely with original baseline performance (cf.~\Cref{fig:overall_accuracy}) and model scale.
Weaker models in the default setting gain between 2.3\% and 5.4\% accuracy, while stronger models lose between 4.8\% and 5.6\%.
As later experiments will further demonstrate, weaker models benefit because the Gym eliminates formatting errors, which caused over 50\% of failures in the baseline setting. %
Stronger models rarely produce invalid paths in the one-shot setting but lose performance in the Gym because the per-step decision frame shifts attention toward immediate moves, limiting global planning even though the full grid remains visible.
This parallels \citet{kambhampati2024llmscantplanhelp, qin2025backtrackbacktracksequentialsearch}, who find that stepwise decomposition helps on locally constrained problems but hurts on tasks requiring long-horizon coordination.

\subsection{Backtracking}
\label{sec:backtracking}

Humans naturally backtrack when solving spatial puzzles, revising previously made decisions when mistakes become apparent \citep{Zamprogno2023AHC}.
In this experiment, we enable models to backtrack, allowing them to undo recent moves.

\begin{figure}[t]
\includegraphics[width=\linewidth]{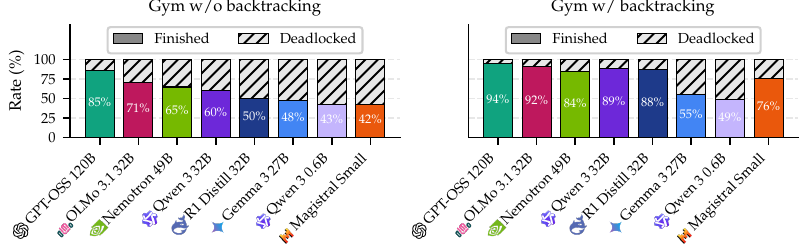}
\vspace{-18pt}
\caption{Completion rates (\%) in Gym w/o backtracking (left) and Gym w/ backtracking (right) for puzzles that were \textit{Finished} (solid) or \textit{Deadlocked} (hatched).}
\label{fig:navigation}
\vspace*{-10pt}
\end{figure}

\Cref{fig:delta}b shows the accuracy delta of models (Backtracking $-$ standard Gym).
The pattern is similar to the formatting-reasoning tradeoff.
Models that performed worse in \Cref{fig:overall_accuracy}, such as Magistral Small, R1 Distill 32B, and Qwen~3 0.6B, show gains of 1.1\% to 2.7\% accuracy, while stronger models, such as GPT-OSS 120B and OLMo 3.1 32B, lose 1.8\% to 5.8\%.
Completion rate improves universally (\Cref{fig:navigation}).
R1 Distill increases completion rate from 50\% to 88\%, Qwen 3 32B from 60\% to 89\%, and GPT-OSS from 85\% to 94\%.
While GPT-OSS finishes more puzzles with backtracking enabled, it solves fewer correctly ($-$5.8\% accuracy).
Reaching the end node is necessary but not sufficient.

Each change from the original one-shot format (Gym w/ or w/o backtracking) results in additional performance loss for the two strongest models, GPT-OSS 120B and OLMo 3.1 32B.
This format sensitivity is consistent with pre-training data contamination, as memorized solution patterns tied to a specific output format break when that format changes \citep{zhang2024carefulexaminationlargelanguage}.
The effect (difference between baseline and Gym w/ backtracking) is strongest for GPT-OSS ($-$11.4\%) and weaker but visible for OLMo ($-$6.6\%).

\begin{wrapfigure}[18]{R}{0.5\linewidth}
\centering
\vspace{-10pt}
\includegraphics[width=\linewidth]{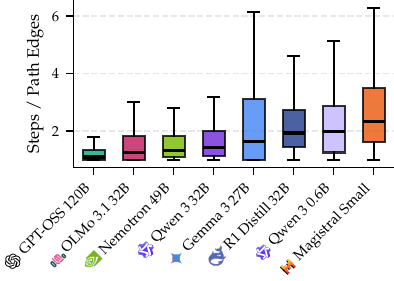}
\vspace{-18pt}
\caption{Backtracking ratio (steps/path edges) per model in \sparcgym{}. Lines show median, boxes show IQR, and whiskers extend to 1.5$\times$ IQR.}
\label{fig:backtracking}
\end{wrapfigure}
\textbf{Backtracking behavior.}
We measure the backtracking frequency as the ratio of the total number of steps to the final path length.
On average, across models, every third action is a backtrack (shown in \Cref{fig:backtracking_scatter} in \Cref{app:backtracking_details}).
\Cref{fig:backtracking} shows the per-model distribution of this ratio.
Backtracking usage correlates inversely with model accuracy.
Magistral Small has the highest median ratio (${\sim}2.2\times$), and GPT-OSS 120B has the lowest (${\sim}1.2\times$).
Models benefiting from backtracking (Magistral Small, R1 Distill, Qwen 3 0.6B, Gemma 3 27B) backtrack heavily, while the four stagnating or collapsing (GPT-OSS, OLMo, Nemotron, Qwen 3 32B) barely do.

Models that performed worse in \Cref{sec:main_results} stumble through heavy backtracking but occasionally find correct solutions they would otherwise miss.
Stronger models, which could benefit most from targeted error correction, rarely backtrack.
This likely reflects training incentives, as RLHF and instruction tuning reward confident responses and penalize self-correction \citep{schulman2017proximalpolicyoptimizationalgorithms}, producing agents that commit to initial decisions even when revision is available.
\citet{qin2025backtrackbacktracksequentialsearch} observe a similar pattern.
Backtracking affects models differently.
Weaker models (see \Cref{fig:overall_accuracy}) improve their completion rate through exploration, while stronger models rarely question initial decisions.

\subsection{Path Generation Effort}
\label{sec:path_effort}
In the baseline evaluation of \citet{kaesberg2025sparcspatialpathfindingreasoning}, humans spent up to $13\times$ longer on hard puzzles than easy ones, while models barely increased their reasoning effort as measured by token usage.
The \sparcgym{} environment makes reasoning efforts observable through path length and step count, in addition to token usage.
\begin{figure}[t]
\includegraphics[width=\linewidth]{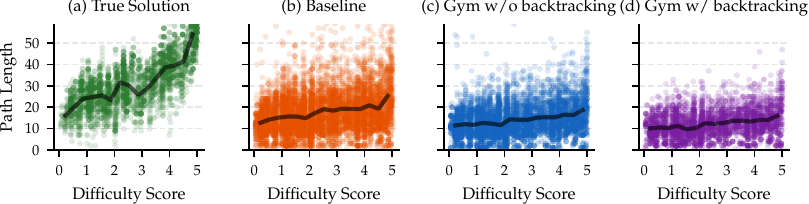}
\vspace{-18pt}
\caption{Path length (in edges) per difficulty score (0--5) for: (a) ground-truth solutions, (b) baseline, (c) \sparcgym{} w/o backtracking, and (d) \sparcgym{} w/ backtracking. Each point represents a single puzzle attempt; the solid line shows the per-difficulty mean.}
\label{fig:path_length}
\end{figure}

\Cref{fig:path_length} compares path length against difficulty across four settings.
(a) Ground-truth solutions show a clear upward trend, as harder puzzles require longer paths.
(b) For the baseline, this upward trend is visible but markedly weaker, with much shorter paths at greater difficulties.
(c) In \sparcgym{}, path lengths compress and barely increase with difficulty, and (d) with backtracking, paths become even shorter, as models use backtracking for pruning rather than to explore alternative routes, the opposite of the intended effect.
We observe that models scale their reasoning tokens up to 5x in the \sparcgym{} over baseline and up to 10x when backtracking is enabled (for details, see \Cref{app:token_usage}).

We observe no correlation between token usage and path length, which shows an inconsistency between reasoning effort and solution quality.
\sparcgym{} forces models to produce more tokens, and this output scales with difficulty.
However, the additional reasoning does not translate into longer or more rule-compliant paths.
Models spend tokens navigating toward shorter paths rather than toward paths that satisfy more constraints, consistent with \citet{snell2024scalingllmtesttimecompute}, who show that more tokens improve performance only when the computation targets the task objective (see \Cref{app:accuracy_tokens}, \Cref{fig:accuracy_vs_tokens} for the token-accuracy mapping). 
We return to this in later experiments (cf. \Cref{sec:ablation_baseline}), where the A* baseline explains why path shortening helps some models but hurts others.
Beyond path lengths and token usage, we track overall accuracy scaling across the five difficulty levels. While the Gym settings yield smoother scaling curves than the one-shot baseline, accuracy universally converges near zero at the highest difficulty levels. This demonstrates a rigid capability limit across all current models, regardless of the interaction format (see \Cref{app:accuracy_difficulty} for difficulty-scaling curves).

\subsection{Rule-Type Analysis}
\label{sec:rule_analysis}
\begin{wrapfigure}[20]{r}{0.5\linewidth}
\vspace{-10pt}
\centering
\includegraphics[width=0.75\linewidth]{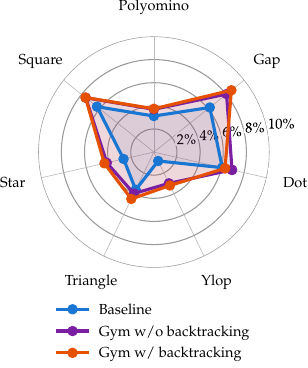}
\vspace{-12pt}
    \caption{Rule-specific accuracy (\%) across all rule types averaged over all models under each of the three evaluation settings.}
\label{fig:radar_average}
\vspace{-10pt}
\end{wrapfigure}

Rule types differ in difficulty, with Gaps being the easiest and Ylops the hardest.
We compute solve rates per rule type across all three settings.
\Cref{fig:radar_average} shows the average across all models.

\sparcgym{} generally helps models improve in all rule types, particularly for Gaps (+2.6\%) and Ylops (+2.3\%).
For some rule types, there is no performance difference between baseline and Gym (e.g., Triangles, Dots).
Backtracking has almost no additional advantage. It helps models reach the end node (\Cref{sec:backtracking}), but does not improve any particular constraint type.

Some improvements are expected as \sparcgym{} eliminates path errors (e.g., Gap rule improvement of 2.6\% over baseline). 
Improvements in Ylops of 2.3\% are interesting because Ylops require shape subtraction across path-created regions, making them the most complex rule type.
In the one-shot setting, models must simultaneously construct a valid path and reason about shape subtraction.
In the Gym setting, this process is iterative, plus models do not need to spend thinking tokens on formatting, freeing capacity for the constraint itself.
The Ylop improvement is a concrete instance of the tradeoff between formatting and reasoning from \Cref{sec:main_results}.
These effects are averaged, and model-specific rule type effects exist, which we provide in \Cref{app:rule_per_model}.

\subsection{Algorithmic Baselines}
\label{sec:ablation_baseline}
Models struggle with navigation and constraint satisfaction, so we compare them against two algorithmic baselines: a random walk (uniform random choice of valid actions) and an A* agent (shortest path from start to end).
The random walk is a lower bound; A* is the ceiling for pure navigation around gaps without rule understanding.
We compare both to the smallest (Qwen 3 0.6B) and largest (GPT-OSS 120B) models.

\begin{figure}[t]
\includegraphics[width=\linewidth]{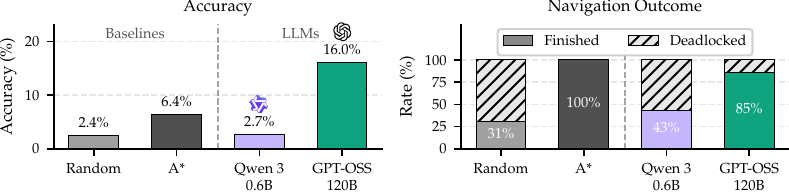}
\vspace{-18pt}
\caption{Accuracy (\%) and completion rate (\%) for random walk, A*, Qwen~3 0.6B, and GPT-OSS 120B in \sparcgym{}. We separate puzzles into \textit{Finished} and \textit{Deadlocked}.}
\vspace*{-12pt}
\label{fig:baselines}
\end{figure}

\Cref{fig:baselines} shows accuracy and completion for the four methods.
The random walk solves 2.4\% of puzzles and completes navigation in 31\% of cases.
A* solves 6.4\% with 100\% completion, showing that pathfinding is not the bottleneck in \sparcgym{}, but constraint satisfaction.
GPT-OSS 120B outperforms A* on accuracy (16.0\% vs.\ 6.4\%) but falls short on navigation (85\% vs.\ 100\%), indicating that LLMs bring rule comprehension that A* lacks at the cost of navigation reliability.
Qwen 3 0.6B solves 2.7\% (43\% completion), marginally above random.

The A* baseline also explains the backtracking results from \Cref{sec:backtracking}.
With backtracking enabled, models optimize toward shorter paths (as seen in \Cref{fig:navigation}), similar to how A* finds the shortest path.
This pulls models toward A*'s accuracy level from both directions.
Models below 6.4\% improve (shorter paths align with A*), while above 6.4\% they lose accuracy (needed detours are removed), explaining why backtracking helps weaker models but hurts stronger ones.

\subsection{Vision}
\label{sec:vision}

\begin{wrapfigure}[15]{r}{0.45\textwidth}
\centering
\vspace{-10pt}
\includegraphics[width=0.8\linewidth]{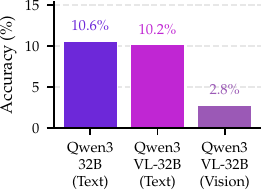}
\vspace{-10pt}
\caption{Vision vs.\ text input for the Qwen3 family. Accuracy (\%) for Qwen3-32B, Qwen3-VL-32B with text input, and Qwen3-VL-32B with puzzle images.}
\label{fig:vision}
\end{wrapfigure}
\leavevmode
Previous work found that multimodal prompting did not improve performance compared to text inputs for one-shot visual puzzles \citet{kaesberg2025sparcspatialpathfindingreasoning}.
We test whether vision-language models can benefit from visual input in the Gym setting.
We compare three conditions within the Qwen~3 family: Qwen3-32B text-only model with text input, Qwen3-VL-32B with text input only, and Qwen3-VL-32B with text and rendered puzzle images.
We use the same puzzle renders shown to human participants.

\Cref{fig:vision} shows the accuracy across these three conditions.
Qwen3-32B (text) and Qwen3-VL-32B (text) perform comparably at 10.6\% and 10.2\%, showing that the vision backbone alone has no impact on text-based spatial reasoning.
Qwen3-VL-32B with vision input collapses to 2.8\%. %
The performance drop is strongest on easier puzzles, where text-based models perform best (\Cref{fig:vision_difficulty} in \Cref{app:vision_details} provides a per-difficulty breakdown).

The vision results presented here are consistent with \citet{park2025generalizingsimplehardvisual}, who find that vision-language models perform notably worse on abstract reasoning tasks when relying on visual inputs compared to equivalent text representations.
When both modalities are provided, the model cannot connect them well and has particular issues indexing the grid system. %

\subsection{Model Scaling}
\label{sec:scaling}
\begin{wrapfigure}[17]{R}{0.5\linewidth}
\centering
\vspace{-18pt}
\includegraphics[width=\linewidth]{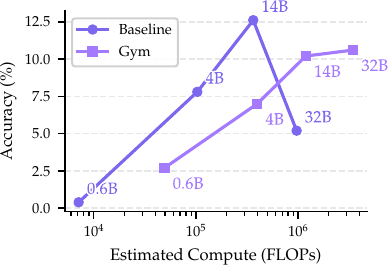}
\vspace{-18pt}
\caption{Accuracy (\%) and estimated compute (FLOPs, log scale) for four Qwen~3 model sizes (0.6B, 4B, 14B, 32B) for the baseline and \sparcgym{}.}
\label{fig:scaling}
\end{wrapfigure}

We evaluate four Qwen~3 model sizes (0.6B, 4B, 14B, 32B) to test whether spatial reasoning scales with model size and whether the two formats produce different scaling trajectories.

\Cref{fig:scaling} shows the results for the baseline and \sparcgym{} tasks.
With the Gym setting, accuracy increases monotonically from 2.6\% at 0.6B to 10.6\% at 32B.
Under the baseline setting, accuracy rises from 0.6B to 14B, peaking at 12.6\%, but the 32B model drops to 5.2\%, falling below the 4B model.
This inversion is specific to the Qwen~3 family. 
Other 32B models (OLMo 3.1 at 16.2\%) and larger models (GPT-OSS 120B at 21.6\%) perform well on the baseline setting.
The Qwen~3 32B model likely produces more formatting errors in the one-shot setting, inflating its failure rate. %
The Gym curve does not show such an artifact.
The 32B model improves over the 14B model's result.
Models are restricted in the Gym setting to choose only among valid moves at each step, rather than constructing a full path; formatting errors are not permitted, and the scaling curve becomes monotonic, reflecting the model's actual spatial capability.
At the same time, the persistently low absolute performance indicates that model scaling alone is insufficient to solve the task. 
We also analyze the behavior of reasoning versus non-reasoning models using Qwen 3 (14B and 32B), finding that extended chain-of-thought retains a $3-5\times$ accuracy advantage (e.g., 10.6\% vs. 2.2\% for 32B) and notably higher completion rates (60-68\% vs. 39-45\%) in the Gym setting (details in \Cref{app:reasoning_mode}).
Motivated by this, we additionally conduct preliminary reinforcement learning experiments, reported in \Cref{app:rl_training}.

\section{Conclusion}
\label{sec:conclusion}

We introduced \sparcgym{}, a Gymnasium-based sequential decision environment for 2D grid pathfinding with interacting rule constraints and optional backtracking.
This sequential decision-making mirrors how humans solve such tasks step-by-step with the ability to backtrack, and addresses the gap in spatial reasoning evaluation by isolating constraint reasoning from output formatting.
We evaluated eight models across three settings (baseline, Gym, Gym+backtracking) against human, random, and A* baselines on 500 puzzles.
The best model, GPT-OSS 120B, achieves 16.0\% accuracy in the Gym setting, 82 points below the human baseline (98.0\%), indicating significant room for improvement in spatial reasoning.
Our results show that the Gym format helps weaker models (up to +5.4\%) by eliminating formatting errors but hurts stronger models (up to $-$5.6\%) by limiting global planning.
Backtracking improves navigation completion universally but only increases accuracy for weaker models that use it heavily, while stronger models rarely backtrack and lose further.
Models also fail to scale reasoning effort with difficulty, as path lengths remain compressed in the Gym setting and backtracking is used to shorten paths rather than explore alternatives.
Algorithmic baselines such as A* show that navigation is largely solved, while constraint satisfaction remains the primary bottleneck in the task.
\sparcgym{} enables RL-based training approaches to directly target the spatial reasoning gap that architecture, scale, and prompting have not closed.

\section{Acknowledgments}

This work was partially supported by the Lower Saxony Ministry of Science and Culture and the VW Foundation. Many thanks to Tobias Mark for his thoughtful discussions and feedback. This work used the Scientific Compute Cluster at GWDG, the joint data center of Max Planck Society for the Advancement of Science (MPG) and University of Göttingen. In part funded by the Deutsche Forschungsgemeinschaft (DFG, German Research Foundation) – 405797229.

\bibliography{colm2026_conference}
\bibliographystyle{colm2026_conference}

\appendix
\crefalias{section}{appendix}%
\crefalias{subsection}{appendix}%
\crefalias{subsubsection}{appendix}%

\section{Models \& Hardware}
\label{app:models_hardware}

\paragraph{Hardware}
All model inference and evaluation procedures for the baseline and \sparcgym{} settings were performed on a uniform hardware configuration. Specifically, we used 4 NVIDIA A100 GPUs, each equipped with 80GB of VRAM. This provided sufficient memory and compute capacity to run all models in a non-quantized mode in our evaluation suite.

\paragraph{Evaluated Models}
We evaluate eight models spanning different sizes, architectures, and training paradigms to assess their spatial reasoning capabilities. The following models were selected for our primary experiments:

\begin{itemize}
    \item \raisebox{-0.1\height}{\includegraphics[width=0.9em]{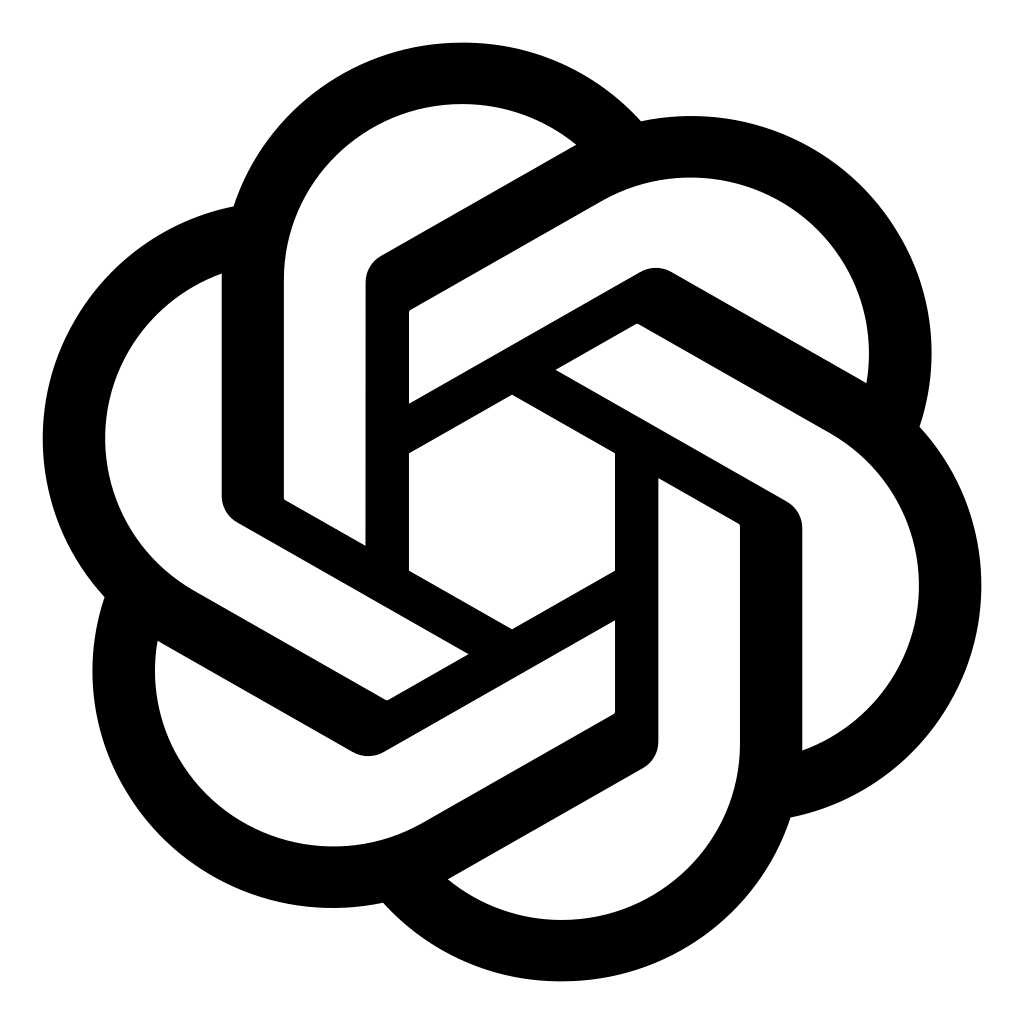}} \textbf{GPT-OSS 120B}: An open-weight, 117B-parameter Mixture-of-Experts (MoE) reasoning model optimized for agentic workflows and high-reasoning tasks \cite{openai2025gptoss120bgptoss20bmodel}.
    \item \raisebox{-0.1\height}{\includegraphics[width=0.9em]{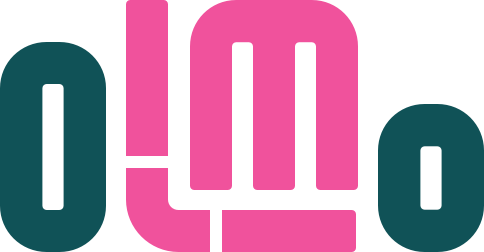}} \textbf{OLMo 3.1 32B}: An open-science language model developed by the Allen Institute for AI, fine-tuned specifically for complex reasoning and multi-step logic \cite{olmo2025olmo3}.
    \item \raisebox{-0.1\height}{\includegraphics[width=0.9em]{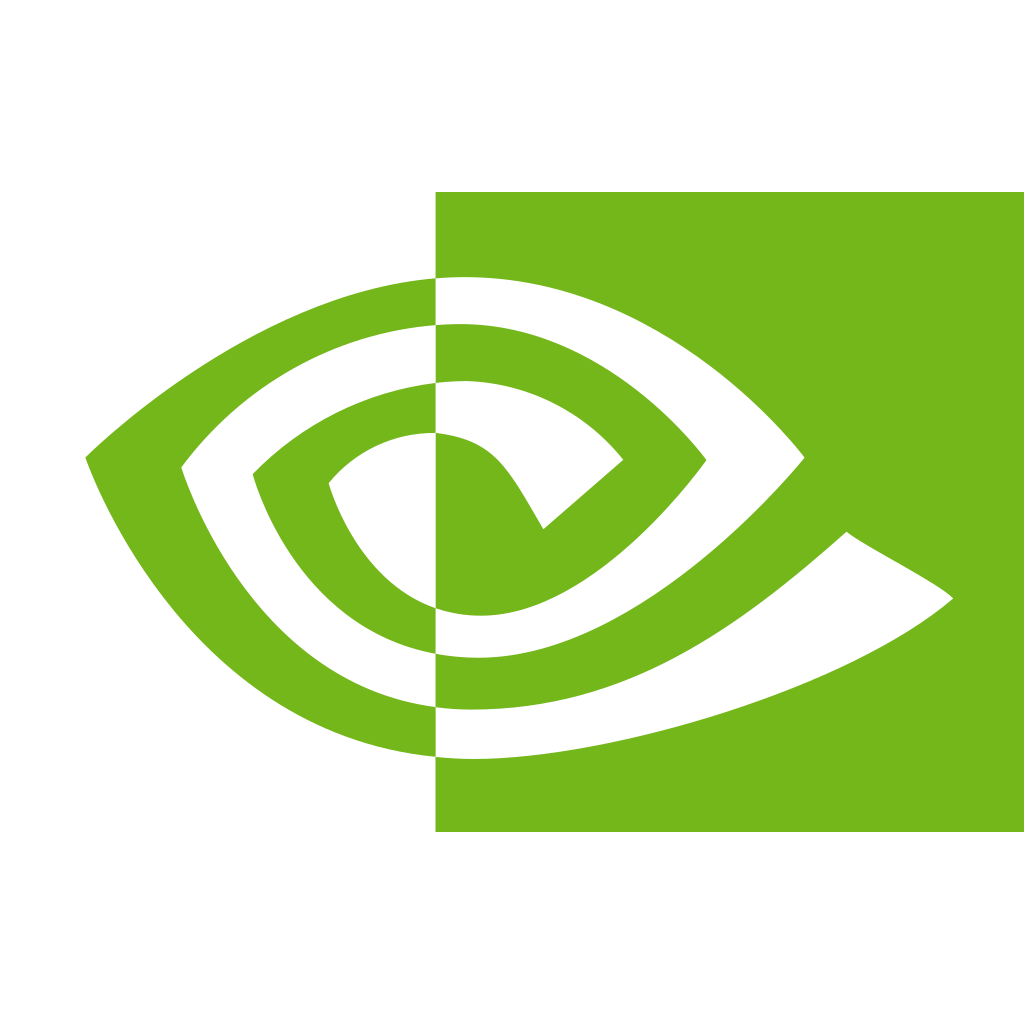}} \textbf{Nemotron 49B}: A highly efficient reasoning derivative (Llama-3.3-Nemotron-Super-49B) developed through Neural Architecture Search \cite{nvidia2025nvidianemotron3efficient}.
    \item \raisebox{-0.1\height}{\includegraphics[width=0.9em]{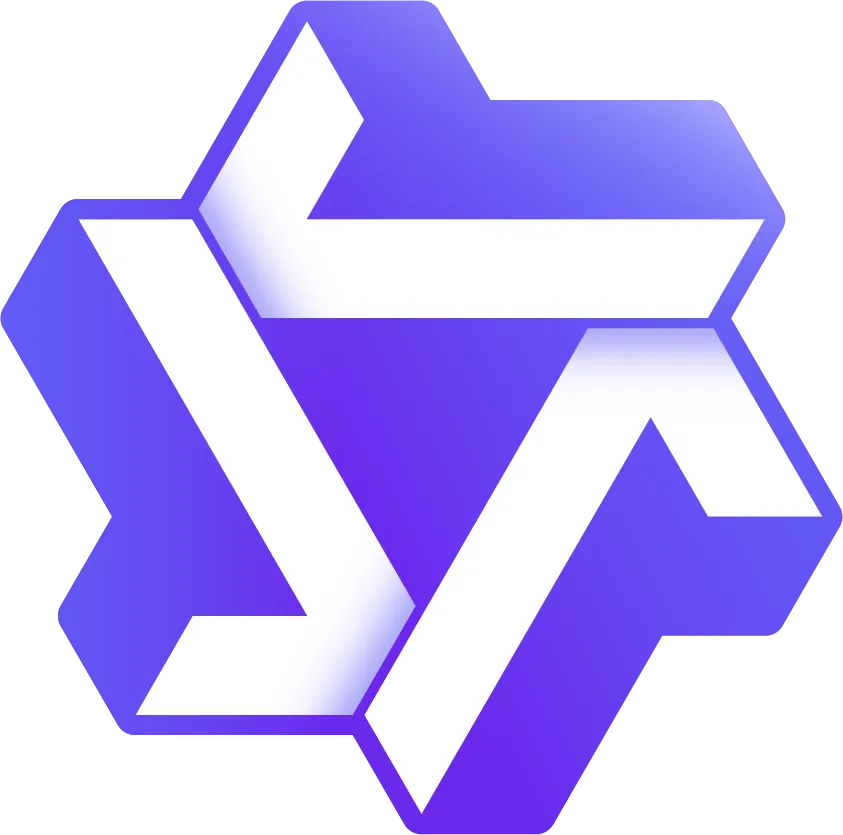}} \textbf{Qwen 3 (32B \& 0.6B)}: Part of Alibaba Cloud's latest model generation featuring native dual thinking and non-thinking modes \cite{yang2025qwen3technicalreport}. Additional variants from this family, including Qwen 3 14B, Qwen 3 4B, and the vision-language model Qwen3-VL-32B, were used specifically for our scaling and vision ablations.
    \item \raisebox{-0.1\height}{\includegraphics[width=0.9em]{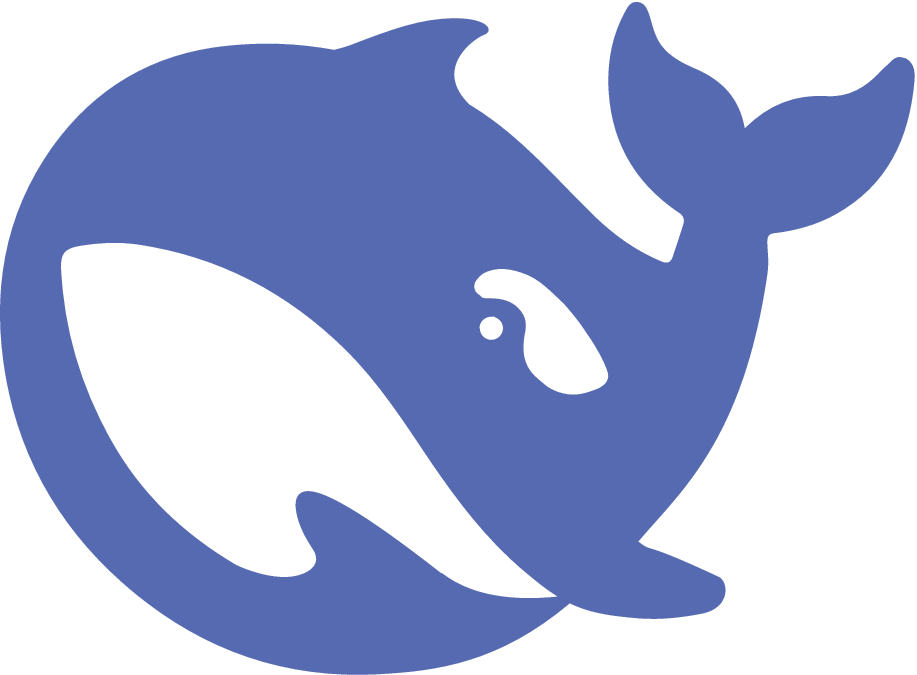}} \textbf{R1 Distill 32B}: DeepSeek-R1-Distill-Qwen-32B, a dense model distilled from the sophisticated reasoning traces of the larger DeepSeek-R1 teacher model \cite{deepseekai2025deepseekr1incentivizingreasoningcapability}.
    \item \raisebox{-0.1\height}{\includegraphics[width=0.9em]{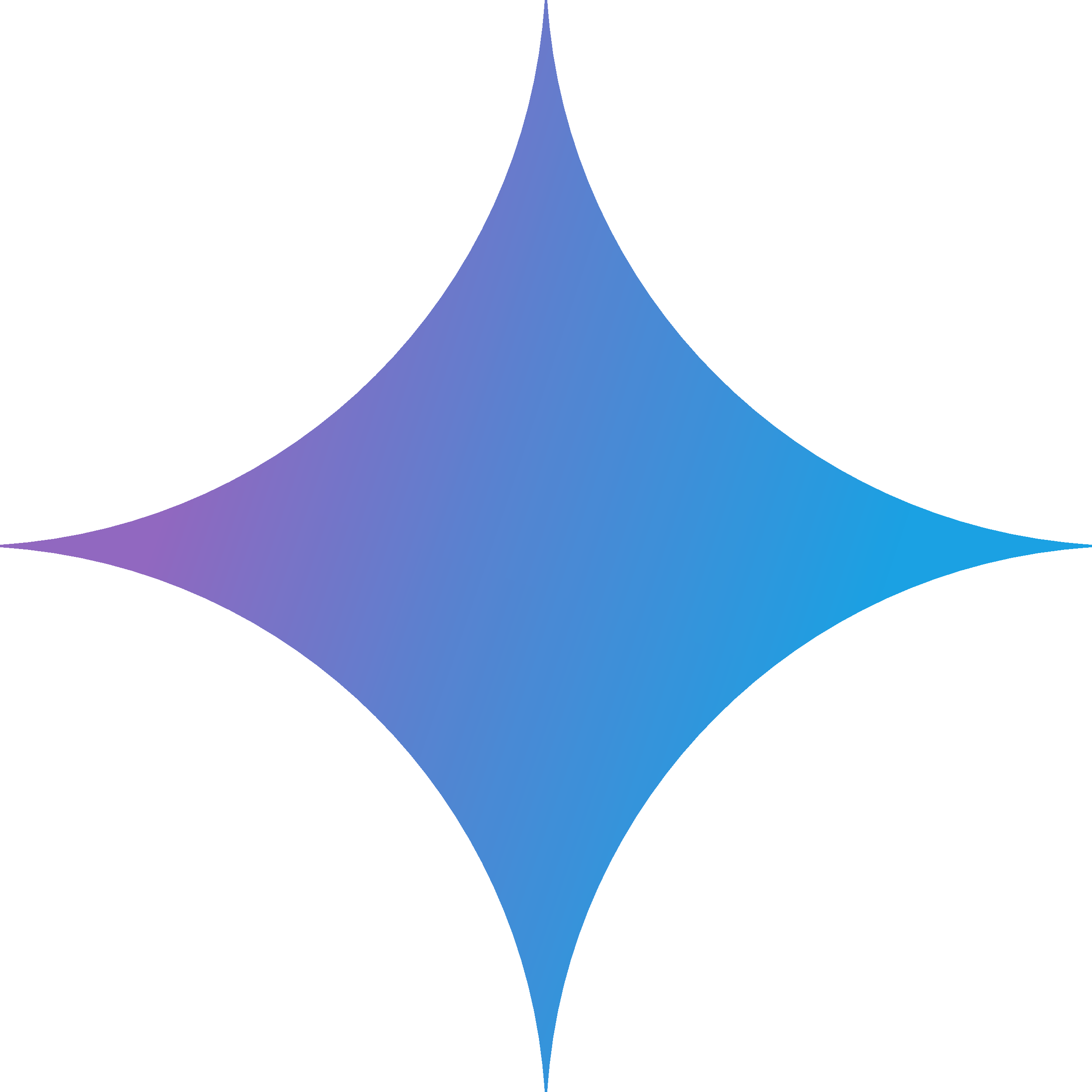}} \textbf{Gemma 3 27B}: Google's multimodal vision-language model, built on Gemini technology and designed for long-context reasoning \cite{gemmateam2025gemma3technicalreport}.
    \item \raisebox{-0.1\height}{\includegraphics[width=0.9em]{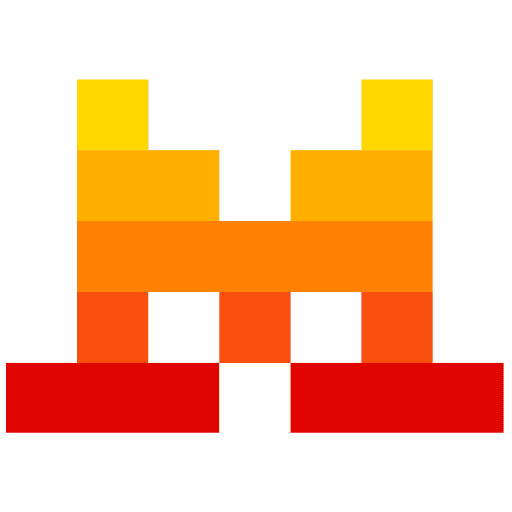}} \textbf{Magistral Small 24B}: A 24-billion parameter reasoning model developed by Mistral AI, fine-tuned for multi-step logic and transparent reasoning traces \cite{mistralai2025magistral}.
\end{itemize}

\section{Prompts}
\label{app:prompts}

This section lists the system prompts used across the three evaluation settings.
All prompts share the same rule definitions and symbol legend; they differ in the instructions for backtracking and input modality.
The user message at each step contains the current observation (grid state, position, legal actions) and is generated dynamically by the environment.
For the visual setting, the user message additionally includes a rendered image of the current puzzle state.

\subsection{Observation Format}
\label{app:prompt-observation}

At each step, the environment generates a user message containing the step number, the agent's current coordinates, the set of legal actions, and the grid state.
The grid encodes the agent's path progress using the symbol legend from the system prompt (\texttt{V} for visited nodes, \texttt{L} for the current position).
The model must end its response with \texttt{Final: <action>}, where \texttt{<action>} is one of the four cardinal directions.
The text-based settings (Spatial-Gym and Spatial-Gym with backtracking) use the template below; the visual setting replaces the grid state with a rendered puzzle image while retaining the position and legal-action text.

\paragraph{User message template.}
\phantom{ }
\begin{lstlisting}[basicstyle=\ttfamily\scriptsize, breaklines=true, frame=single, columns=fullflexible, keepspaces=true, xleftmargin=0.5em, xrightmargin=0.5em]
Step: {step}
Current Position: ({x}, {y})
Legal Actions: [{digit=DIRECTION, ...}]

Grid State:
{observation grid}

You MAY think step-by-step, but you MUST end your response with:
Final: <action>
Where <action> is one of 0=RIGHT, 1=UP, 2=LEFT, 3=DOWN
  (you may also write the direction name, e.g. Final: right).
\end{lstlisting}

\paragraph{Example observation (step~1).}
\phantom{ }
The following shows the first observation for a difficulty-5 puzzle with a 3$\times$2 grid containing a polyshape, two blue squares, and a dot.
The agent starts at position $(0,1)$, marked \texttt{L}.

\begin{lstlisting}[basicstyle=\ttfamily\scriptsize, breaklines=true, frame=single, columns=fullflexible, keepspaces=true, xleftmargin=0.5em, xrightmargin=0.5em]
Step: 1
Current Position: (0, 1)
Legal Actions: [1=UP,3=DOWN]

Grid State:
['+', '+', '+', '+', '+', '.', '+']
['L', 'N', '+', 'P-R-16', '+', 'o-B', '+']
['+', '+', '+', '+', '+', '+', '+']
['+', 'N', '+', 'N', '+', 'o-B', '+']
['+', '+', 'E', '+', '+', '+', '+']

You MAY think step-by-step, but you MUST end your response with:
Final: <action>
Where <action> is one of 0=RIGHT, 1=UP, 2=LEFT, 3=DOWN
  (you may also write the direction name, e.g. Final: right).
\end{lstlisting}

\subsection{Spatial-Gym (Step-by-Step without backtracking)}
\label{app:prompt-gym}

\begin{lstlisting}[basicstyle=\ttfamily\scriptsize, breaklines=true, frame=single, columns=fullflexible, keepspaces=true, xleftmargin=0.5em, xrightmargin=0.5em]
You are an autonomous agent controlling a path-finding puzzle solver.
Your goal is to find a valid path (a continuous line) from the specified
Start Node to the End Node on the provided grid, adhering to all puzzle
rules.

Core Concepts & Grid Basics:
Grid Dimensions: You can find the puzzle grid size in the info
Path: The solution is a single, continuous line connecting adjacent nodes
  either horizontally or vertically.
Revisiting: You can not traceback your path. You can not visit a cell
  twice.
Rule Cells: Cells containing rule symbols (squares, stars, etc.) have
  coordinates where both x and y are odd. The path goes around these rule
  cells, never on them. They are also marked as gaps.
Regions: The drawn path divides the grid cells into one or more distinct
  enclosed areas (regions). Many rules apply based on the contents of
  these regions.
Valid Path Cells: The path travels along the grid lines (edges between
  nodes). It can only occupy positions marked '+' or '.' in the grid
  layout (these correspond to positions with at least one even
  coordinate).

Symbol Legend (Grid Notation)
  S: Start Node (Path begins here)
  E: End Node (Path ends here)
  V: Visited Node (Path has passed through this cell)
  L: Current Node (Path is currently on this cell)
  +: Valid cell for the path to occupy
  N: Empty rule cell (no rule)
  G: Gap (Path CANNOT cross this cell)
  .: Dot (Path MUST pass through this cell)
  o-X: Square of color X
  *-X: Star of color X
  A-X: Triangle (touch 1 edge)
  B-X: Triangle (touch 2 edges)
  C-X: Triangle (touch 3 edges)
  D-X: Triangle (touch 4 edges)
  P-X-Y: Polyshape (positive) of color X and shape ID Y
  Y-X-Y: Negative Polyshape (ylop) of color X and shape ID Y

Color Codes: R=Red, B=Blue, G=Green, Y=Yellow, W=White, O=Orange,
  P=Purple, K=Black

Detailed Solving Rules:
The drawn path must satisfy ALL applicable constraints:

1. Path Constraints:
   Path connects adjacent nodes (horizontal/vertical moves only).
   Nodes CAN NOT be revisited. You cannot visit a cell twice.
   Path MUST pass through all Dot cells.
   Path CANNOT pass through any Gap cells.

2. Region-Based Rules (Apply to areas enclosed by the path):
   Squares: All squares within a single region MUST be the same color.
     Squares of different colors MUST be separated into different regions
     by the path.
   Stars: Within a single region, each star symbol MUST be paired with
     exactly ONE other element of the same color. Other colors within the
     region are irrelevant to this specific star's rule.
   Polyshapes (poly): The region containing this symbol MUST be able to
     contain the specified shape (defined in Polyshape Definitions). The
     shape must fit entirely within the region's boundaries. If multiple
     positive polyshapes are in one region, the region must accommodate
     their combined, non-overlapping forms. Rotation of polyshapes is NOT
     allowed. They must fit within the provided space in their given
     orientation.
   Negative Polyshapes (ylop): These subtract shape requirements,
     typically within the same region as corresponding positive
     polyshapes. A negative polyshape cancels out a positive polyshape of
     the exact same shape and color within that region. If all positive
     shapes are canceled, the region has no shape constraint.

3. Path-Based Rules (Edge Touching):
   Triangles: The path MUST touch a specific number of edges of the cell
     containing the triangle symbol.
     (1): Path touches EXACTLY 1 edge of the triangle's cell.
     (2): Path touches EXACTLY 2 edges of the triangle's cell.
     (3): Path touches EXACTLY 3 edges of the triangle's cell.
     (4): Path touches EXACTLY 4 edges (fully surrounds) the cell.

Polyshape Definitions: Shapes are defined by 2D arrays where 1 indicates
  an occupied cell and 0 indicates an empty cell.
{polyshapes}

At each turn you'll receive the current state:
- Step: The current step number
- Current Position: Your current (x, y) location
- Legal Actions: Available moves with format [digit=DIRECTION, ...]
- Grid State: The current grid showing your path progress
\end{lstlisting}

\subsection{Spatial-Gym with backtracking}
\label{app:prompt-traceback}

The backtracking prompt is identical to the standard Gym prompt (Appendix~\ref{app:prompt-gym}) except for the revisiting policy and the corresponding path constraint.
All other sections (symbol legend, solving rules, polyshape definitions, observation format) remain unchanged.
Below we list only the two segments that differ.

\paragraph{Core Concepts (Revisiting).}
\phantom{ }
\begin{lstlisting}[basicstyle=\ttfamily\scriptsize, breaklines=true, frame=single, columns=fullflexible, keepspaces=true, xleftmargin=0.5em, xrightmargin=0.5em]
Revisiting: You can traceback your path, but you MUST do so in the same
  way you came, without crossing over your own path. When tracing back,
  you can only move to the last cell you occupied, and then continue from
  there. Also when you traceback, the nodes you no longer use in your
  path are free to be used again.
\end{lstlisting}

\paragraph{Path Constraints.}
\phantom{ }
\begin{lstlisting}[basicstyle=\ttfamily\scriptsize, breaklines=true, frame=single, columns=fullflexible, keepspaces=true, xleftmargin=0.5em, xrightmargin=0.5em]
1. Path Constraints:
   Path connects adjacent nodes (horizontal/vertical moves only).
   Nodes CAN be revisited. But only if you traceback to the last cell
     you occupied (and from there again and again ...). Otherwise you
     CANNOT cross your own path.
   Path MUST pass through all Dot cells.
   Path CANNOT pass through any Gap cells.
\end{lstlisting}

\subsection{Spatial-Gym Visual}
\label{app:prompt-visual}

The visual prompt replaces the text-based grid representation with a rendered image of the puzzle state at each step.
The model receives both the image and a text message containing coordinates and legal actions.
The system prompt describes the visual appearance of all puzzle elements so the model can ground symbols in the image.

\begin{lstlisting}[basicstyle=\ttfamily\scriptsize, breaklines=true, frame=single, columns=fullflexible, keepspaces=true, xleftmargin=0.5em, xrightmargin=0.5em]
You are an autonomous agent controlling a path-finding puzzle solver.
Your goal is to find a valid path (a continuous line) from the specified
Start Node to the End Node on the provided grid, adhering to all puzzle
rules.

## Input Format
At each step you receive TWO inputs:
1. An image of the puzzle showing the current state, including your path
   progress
2. A text message with your current position coordinates and available
   legal moves

Use both together -- the image helps with spatial reasoning and visual
pattern recognition, while the text provides precise position and action
information.

## Visual Appearance of the Puzzle Image

The image shows a Witness puzzle grid with a dark teal/green background.
Here is how elements appear visually:

Board structure:
- Teal/green cells separated by dark gray grid lines (the path network)

Navigation markers:
- Start Node: A large filled circle on the grid edge -- this is where the
    path begins
- End Node: A small rounded nub/extension protruding outward from the
    grid edge -- this is where you must reach
- Visited Path: Marked with a WHITE LINE showing where you have already
    traveled
- Current Position: Located at the END of the white line (where the line
    stops)

Rule symbols in cells (located in cells where both x and y are odd --
  the path goes AROUND these cells, not through them):
- Colored Squares: Colored rounded rectangles inside cells (e.g., black,
    red, blue)
- Colored Stars: 8-pointed star shapes in their respective color
- Triangles: Small colored upward-pointing triangles inside cells -- the
    count (1-4 triangles) indicates how many edges of that cell the path
    must touch
- Polyshapes (positive): Tetromino-like colored filled block patterns
    inside cells, showing the shape the region must match
- Polyshapes (negative/ylop): Same block patterns but drawn as
    hollow/outlined squares instead of filled -- these cancel out
    positive polyshapes

Path elements (located on grid lines):
- Dots: Small black hexagons on the grid lines -- the path MUST pass
    through these
- Gaps: Broken/missing segments on the grid lines -- the path CANNOT
    cross these

## Coordinate System
- Nodes are indexed (x, y) where (0,0) is the top-left node
- x increases to the right, y increases downward
- The path travels along grid edges, connecting adjacent nodes
    horizontally or vertically

## Solving Rules
1. Draw a continuous line from START to END without visiting the same
   node twice
2. The line can only be placed on valid path cells (not on rule cells)
3. The line acts as a boundary, dividing the grid into regions
4. All rule symbols must be satisfied:
   - Dots (black hexagons on grid lines): The line MUST pass through each
       dot
   - Colored Squares (filled rounded rectangles in cells): All squares in
       a single region must be the same color. Different colors MUST be
       separated into different regions
   - Colored Stars (8-pointed stars in cells): Each star must be paired
       with EXACTLY one other element of the same color in its region
   - Triangles (small triangles in cells): The line must touch EXACTLY
       the number of edges specified by the triangle count (1-4 edges)
   - Polyshapes (filled block patterns in cells): The region must be
       shaped exactly like the defined polyshape
   - Negative Polyshapes (hollow block patterns in cells): Cancel out
       regular polyshapes if they overlap

{polyshapes}

## Actions
You can move in 4 directions:
- 0: Move UP
- 1: Move RIGHT
- 2: Move DOWN
- 3: Move LEFT

## Important
- You CANNOT revisit cells you have already visited
- You CANNOT cross your own path
- Plan ahead to satisfy all constraints

## How to Respond
Analyze the puzzle image showing your current position and path progress,
together with the text information about your coordinates and legal
moves. Determine which direction leads toward the exit while satisfying
all puzzle rules.
\end{lstlisting}

\section{SPaRC Dataset}
\label{app:sparc_dataset}

SPaRC~\citep{kaesberg2025sparcspatialpathfindingreasoning} consists of 1{,}000 2D grid pathfinding puzzles (500 train, 500 test) inspired by the puzzle mechanics of \textit{The Witness}~\citep{witness}.
Each puzzle is an $m \times n$ grid of \textbf{rule cells} with $(x,y)=(0,0)$ at the top-left corner, $x$ increasing rightward and $y$ downward.
Rule cells are surrounded by \textbf{edges} along which the solution path is drawn.
Each puzzle has exactly one \textbf{start point} (large circle) and one \textbf{end point} (edge extension).
The goal is to draw a single, continuous, non-self-intersecting path along edges from start to end that satisfies all rule cell constraints.

\subsection{Rules}
\label{app:sparc_rules}

Seven rule types can appear in SPaRC puzzles:

\long\def\figwindownonum[#1,#2,#3,#4] {%
  \begin{window}[#1,#2,{#3},{\centering#4\par}] }
\def\endfigwindownonum{\end{window}}%

\newcommand{\iconheight}{5ex}

\newcommand{\puzzleitem}[3]{%
  \par\smallskip                    %
  \begin{figwindownonum}[0,l,{
    \includegraphics[width=\iconheight]{#1}},{}]
    \noindent\textbf{#2:} #3%
  \end{figwindownonum}%
  \par\medskip                     %
}

\puzzleitem{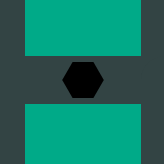}{Item Collection (Dots)}{%
The solution path needs to pass through every dot.
}

\puzzleitem{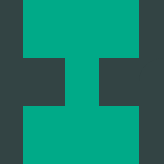}{Path Breaks (Gaps)}{%
The solution path cannot go through any edge segment containing a gap. Gaps act as local barriers.
}

\puzzleitem{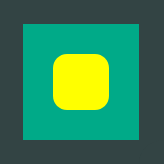}{Color Separation (Stones)}{%
The solution path must be drawn to separate stones of different colors. All stones located within any single enclosed region must be of the same color.
}

\puzzleitem{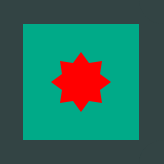}{Pairing (Stars)}{%
Each star must share its region with exactly one other symbol of the same color. No unpaired stars are allowed.
}

\puzzleitem{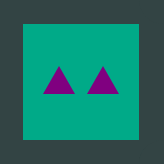}{Edge Count (Triangles)}{%
The solution path must touch the number of edges shown by the triangles in the cell, e.g., two triangles mean the path must touch exactly two edges of that cell.
}

\puzzleitem{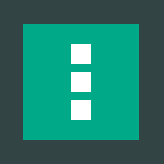}{Shape Fitting (Polyominoes)}{%
If a cell contains a polyomino (poly), the solution path must enclose a region that matches its exact shape and area. The region must not rotate or mirror the poly. Multiple polys can share a region if their shapes fit without overlapping.
}

\puzzleitem{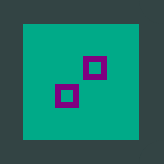}{Shape Subtraction (Ylop)}{%
A ylop must be enclosed in the same region as one or more polys. Its shape and area subtract from the total required by the polys. If a ylop cancels out a poly exactly, that pair imposes no constraint.
}

\subsection{Dataset Creation}
\label{app:dataset_creation}
 
Puzzles are generated by randomly creating an $m \times n$ grid ($m, n \in [2, 6]$), filling approximately half the cells with rules (\textit{rule density}), and placing random start and end points.
A generation-validation loop solves each puzzle via brute-force search over all valid paths.\footnote{Brute-force is necessary because many puzzles fall into NP or NP-Complete complexity classes~\citep{abel2019witnesseswitnessfindingwitnesses}.}
If no solution exists, rule density is decreased; if more than $k{=}50$ solutions exist, density is increased, and the puzzle is regenerated.
 
\paragraph{Difficulty estimation.} Puzzle complexity is quantified by a weighted sum of the number of distinct rule types, total rule cells, rule density, grid size, and estimated rule interactions, normalized to a 1 (easiest) to 5 (hardest) scale.
The test set contains 86 puzzles at level~1, 118 at level~2, 121 at level~3, 86 at level~4, and 89 at level~5.

\section{Example Gym Interaction}
\label{app:example_gym}

\noindent\small
\begin{tabular}{@{}p{0.22\linewidth}p{0.72\linewidth}@{}}
\textbf{Puzzle ID} & c2f1726c32030b96 \\
\textbf{Difficulty} & GPT-OSS 120B \\
\textbf{Difficulty} & 2 (score 1.22) \\
\textbf{Grid (H$\times$W)} & 2$\times$2 \\
\textbf{Outcome} & Solved \\
\end{tabular}
\par\vspace{1em}
\noindent\rule{\linewidth}{0.4pt}\par\medskip
\noindent\textbf{Step 1 of 5}
\par\smallskip
\noindent\textbf{Chosen action:} \textbf{Down}
\par
\begin{wrapfigure}[10]{l}{0.26\textwidth}
\centering
\vspace{-12pt}
\includegraphics[width=\linewidth]{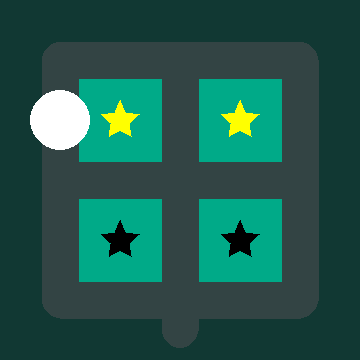}
\end{wrapfigure}
\noindent\textbf{Model reasoning}\par\smallskip
\noindent{\raggedright\footnotesize\ttfamily We need to analyze puzzle. Grid is 5x5? Represented as rows lines. Coordinates? Current Position: [0 1] means x=0, y=1? Usually x column, y row. Starting at coordinate (0,1). Let's map grid with coordinates (x from 0 left to 4, y from 0 top to 4). Grid rows given:\par Row0: ['+', '+', '+', '+', '+']\newline Row1: ['S', '*-Y', '+', '*-Y', '+']\newline Row2: ['+', '+', '+', '+', '+']\newline Row3: ['+', '*-K', '+', '*-K', '+']\newline Row4: ['+', '+', 'E', '+', '+']\par Symbols: S at (0,1). E at (2,4). stars at (1,1) color Y, (3,1) Y, (1,3) K, (3,3) K.\par Current Position is [0 1] which matches S. Legal actions: up or down. Means cannot go left or right because maybe out of bounds? From (0,1), up goes to (0,0) which is '+'. Down goes to (0,2) which is '+'.\par Goal: Find a path to E covering all dot cells (none) and obey triangle (none) etc. Rule: cannot pass through rule cells (stars) since they are gaps? Stars are rule cells at odd coordinates (both odd). Indeed coordinates (1,1), (3,1), (1,3), (3,3) are rule cells; path cannot go through them.\par We need to find a path from S to E without crossing those star cells, cannot revisit, and must end at E (2,4). Also there may be region rules: squares none, triangles none, poly shapes none. Only stars: each star must be paired with exactly one other element of same color within its region. Stars of color Y at (1,1) and (3,1). Stars of color K at (1,3) and (3,3). They are each color with two stars; they need to be paired with exactly one other element of same color inside same region. The other element could be maybe another star? Actually "each star symbol MUST be paired with exactly ONE other element of the same color." Could be another star? Probably star pairs themselves count? Not sure. Usually in puzzles like 'Star Pairing', each star must be paired with a dot of same color. But here maybe another element (could be dot, square, etc.). No other same color elements exist besides the stars themselves. So each star must be paired with the other star of same color, but they are separate stars. "Each star symbol must be paired with exactly ONE other element of the same color. Other colors within region are irrelevant." So if there are two Y stars in same region, they could pair with each other; each star paired exactly one other element (the other star). That satisfies rule if they are in same region. But if they are separated by path into different regions, they'd each lack a partner. So we need to ensure that stars of same color are in same region. So the path must not separate Y stars from each other, same for K stars.\par Thus we need to draw path from S to E that encloses a region containing both Y stars together, and another region containing both K stars together, or maybe they can be in same region as each other? But if Y star paired with K star? No, must be same color, so not.\par Simplify: we can make a path that goes around all four star cells maybe creating a region that contains them all together (both colors). That would put Y stars with K stars together, but each Y star still has partner? It could be the other Y star in same region (still there). So okay.\par But path cannot go through them, can go on surrounding '+' cells.\par We need to find any path from S (0,1) to E (2,4) that doesn't cross star cells and doesn't revisit. Let's attempt to design moves.\par Grid visualize with coordinates (x,y):\par (0,0) + ; (1,0)+ ; (2,0)+ ; (3,0)+ ; (4,0)+\newline (0,1) S ; (1,1)*-Y ; (2,1)+ ; (3,1)*-Y ; (4,1)+\newline (0,2)+ ; (1,2)+ ; (2,2)+ ; (3,2)+ ; (4,2)+\newline (0,3)+ ; (1,3)*-K ; (2,3)+ ; (3,3)*-K ; (4,3)+\newline (0,4)+ ; (1,4)+ ; (2,4)E ; (3,4)+ ; (4,4)+\par We can move through any '+' cells.\par We start at (0,1). Allowed moves up to (0,0) or down to (0,2). Need to eventually go to (2,4). Let's think simple path: go down, right, down, right, down, right? But must avoid stars at (1,3) and (3,3). Also (1,1) and (3,1) are stars.\par Potential path: from S down to (0,2), then down to (0,3), then down to (0,4), then right to (1,4), right to (2,4) reaching E. This path would be: S (0,1) -> (0,2) -> (0,3) -> (0,4) -> (1,4) -> (2,4) E. This does not cross any star cells. Does it satisfy star pairing region rules? The path will create only one region (outside) and maybe interior region? The path is a simple line that does not enclose any area because it's not closed loop; path divides plane but region concept maybe only closed loops create enclosed areas. But stars are not in the path, they are on the other side of path. For star pairing, they need to be in same region; path does not partition them separate as they are on same side (since path is on left side of them). All stars are to the right of path (since path stays at x<=0 then x=1 at (1,4) but still left of stars). Since they stay together, they are in same region (the region not occupied by path). So they are paired within same region. So rule satisfied.\par Thus this simple path seems valid. No dot constraints.\par Thus first move: we must choose from legal actions: up or down. Down leads to (0,2) which continues as described. Up would go to (0,0) maybe also possible but then need to navigate. Both may lead to solution.\par Pick down (3). So final answer digit 3.\par We must output "Final: 3".\par}
\par\bigskip
\noindent\rule{\linewidth}{0.4pt}\par\medskip
\noindent\textbf{Step 2 of 5} 
\par\smallskip
\noindent\textbf{Chosen action:} \textbf{Down}
\par
\begin{wrapfigure}[9]{l}{0.26\textwidth}
\centering
\vspace{-12pt}
\includegraphics[width=\linewidth]{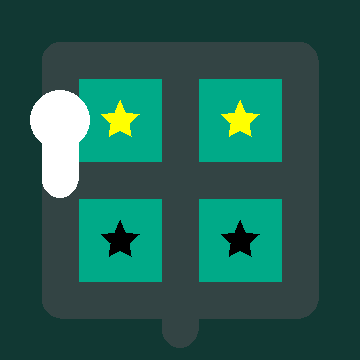}
\end{wrapfigure}
\noindent\textbf{Model reasoning}\par\smallskip
\noindent{\raggedright\footnotesize\ttfamily We need to decide next move. Let's interpret grid. Coordinates? Current position [0 2] means x=0, y=2? Grid indexing probably row,column? Given grid displayed as rows. Row0: ['+', '+', '+', '+', '+']\newline Row1: ['V', '*-Y', '+', '*-Y', '+']\newline Row2: ['L', '+', '+', '+', '+']\newline Row3: ['+', '*-K', '+', '*-K', '+']\newline Row4: ['+', '+', 'E', '+', '+']\par Current position L at row2 col0. Legal actions: [0=RIGHT, 3=DOWN]. So can go right to col1 row2 or down to row3 col0.\par Goal is to reach End E at row4 col2. Must also pass through any dots '.' (none visible). Must avoid gaps G (none). Must go around rule cells squares, stars etc. We have stars at positions (row1 col1) and (row1 col3) color Y. Also stars at (row3 col1) and (row3 col3) color K. Triangles? none. Squares? none. Need to satisfy region constraints: each region squares same color; not relevant. Stars: each star must be paired with exactly one other element of same color within same region. That element can be? Probably another star of same color? Or maybe something else? Rule says each star must be paired with exactly ONE other element of same color. Other colors irrelevant. Likely need to have stars of same color paired together within same region. So Y stars must be paired together, K stars together. Must be in same region (enclosed area separated by path). So path must separate Y and K stars maybe each region contains both Y stars? Actually they are currently separate cells: Y stars at (1,1) and (1,3). K stars at (3,1) and (3,3). Path can separate them into regions such that each region contains exactly two stars of same color perhaps? The rule: each star must be paired with exactly one other element of same color. So each Y star must pair with another Y (the other Y star). So both Y stars should be in same region. Similarly K stars in same region. Therefore path must not separate Y stars from each other; same for K.\par Currently path started at some start node at maybe top-left? We have visited cell V at row1 col0, indicating path from start has moved down? Let's track: Starting location? Not given, but currently visited V at row1 col0. L at row2 col0. So path goes from start somewhere above row0 maybe? Probably start at (0,0) top-left? Actually grid top-left row0 col0 is '+'. Possibly start S is at that location but not shown because visited? Wait V at row1 col0 indicates that cell was visited earlier. So start maybe at row0 col0? Then moved down to row1 col0 (now V), then down to row2 col0 (now L). So path downwards column0.\par Goal is to reach E at (4,2). Need to navigate around stars. Since stars are at (1,1) (1,3) (3,1) (3,3). Path cannot go through them because they are rule cells (odd coordinates both). Indeed they are '*' entries. Path can go around them.\par Current path column0. Options: Right to column1 row2, which would move into a cell currently '+' at row2 col1. That cell is adjacent to Y star at (1,1) (above) and K star at (3,1) (below). Passing through there will affect region boundaries.\par Alternatively down to row3 col0 which is '+' (currently). Let's consider overall plan: need to end at (4,2). Likely path will go down to bottom then right to column2. Must also ensure stars are paired properly: both Y stars must be in same region. If we draw a path that separates top row Y stars from each other? They are separated by column2 (col2 is '+'), but they are at col1 and col3 with a plus cell between. Path can go between them to separate them into different regions. To keep them together, we need to ensure path does not cut between them. Similarly for K stars.\par Thus, we should avoid creating a wall that cuts between the two Y stars horizontally across row1 between col1 and col3. That would be at row1 col2, which is currently '+'. If path passes through row1 col2, it would separate Y stars into different regions (above? Actually path goes through that cell; the region on one side might separate). Hard to reason.\par Given start column0, maybe we go down to row3 col0 then right across bottom row, then up etc. Let's sketch potential path: currently at (2,0). Down to (3,0), then down to (4,0), then right to (4,1), right to (4,2) which is E. That would be straightforward, not intersecting any stars. Does any rule forbid passing adjacent to stars? No.\par Check legal actions: after moving down from (2,0) to (3,0), would there be legal moves? Need to see if path passes through any '.' none.\par Thus probably best move is DOWN (digit 3). Let's ensure not revisiting: (3,0) is not visited yet. No constraints broken.\par Thus final decision digit 3.\par}
\par\bigskip
\noindent\rule{\linewidth}{0.4pt}\par\medskip
\noindent\textbf{Step 3 of 5} 
\par\smallskip
\noindent\textbf{Chosen action:} \textbf{Down}
\par
\begin{wrapfigure}[10]{l}{0.26\textwidth}
\centering
\vspace{-12pt}
\includegraphics[width=\linewidth]{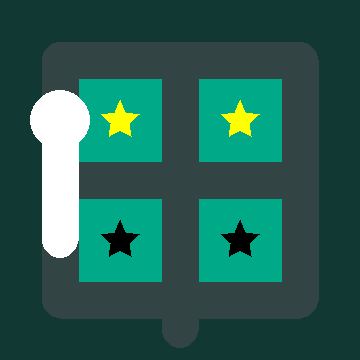}
\end{wrapfigure}
\noindent\textbf{Model reasoning}\par\smallskip
\noindent{\raggedright\footnotesize\ttfamily We have a path-finding puzzle. Need to decide next move. Step 3, current position [0 3]. Coordinates likely x y? Grid representation: rows? They show grid as 5x5. Indices maybe [x y] means column x, row y? Starting at (0,3) which is first column, fourth row (0-index). Grid shows row order: first row ['+', '+', '+', '+', '+']; second row ['V', '*-Y', '+', '*-Y', '+']; third row ['V', '+', '+', '+', '+']; fourth row ['L', '*-K', '+', '*-K', '+']; fifth row ['+', '+', 'E', '+', '+'].\par At current position L at row4 column0 (0-index). Adjacent moves: Right would go to '*-K' (a star) which is a rule cell (odd coordinates) maybe not allowed for path (path cannot go onto rule cells). According to core concept: Path travels only on positions marked '+' or '.' . In grid, '*' cells are rule cells - path can't go on them. So right is not legal. Up would go to row3 column0 which is 'V' (visited). Cannot revisit? Path cannot revisit cells. So up not legal. Left would go out of bounds. Down goes to row5 column0 which is '+' (bottom row). Legal actions list only 3=DOWN. So must go down. So final answer digit 3.\par}
\par\bigskip
\noindent\rule{\linewidth}{0.4pt}\par\medskip
\newpage
\noindent\textbf{Step 4 of 5} 
\par\smallskip
\noindent\textbf{Chosen action:} \textbf{Right}
\par
\begin{wrapfigure}[10]{l}{0.26\textwidth}
\centering
\vspace{-12pt}
\includegraphics[width=\linewidth]{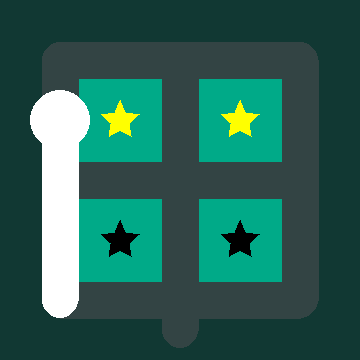}
\end{wrapfigure}
\noindent\textbf{Model reasoning}\par\smallskip
\noindent{\raggedright\footnotesize\ttfamily We need to decide the move. Let's understand grid layout. Coordinates: current position [0 4] meaning x=0, y=4? Probably (col, row) with origin top-left (0,0). Grid shown rows from top (row0) to bottom (row4). Row0: ['+','+','+','+','+'] columns 0-4.\par Row1: ['V','*-Y','+','*-Y','+']\newline Row2: ['V','+','+','+','+']\newline Row3: ['V','*-K','+','*-K','+']\newline Row4: ['L','+','E','+','+']\par Current position L at row4 col0. End E at row4 col2. There's a plus at row4 col1 between L and E. Also gap? no.\par Legal actions: [0=RIGHT] only. So must move right to column1 row4. That's allowed.\par Check other constraints: path cannot revisit, and must pass through all dots (none). Should be okay.\par Thus final action 0.\par But ensure cannot go through gap? The cell at row4 col1 is '+', not gap.\par Thus answer.\par}
\par\bigskip
\noindent\rule{\linewidth}{0.4pt}\par\medskip
\noindent\textbf{Step 5 of 5} 
\par\smallskip
\noindent\textbf{Chosen action:} \textbf{Right}
\par
\begin{wrapfigure}[10]{l}{0.26\textwidth}
\centering
\vspace{-12pt}
\includegraphics[width=\linewidth]{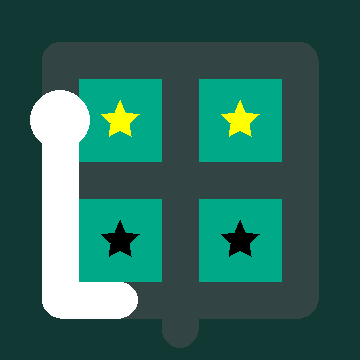}
\end{wrapfigure}
\noindent\textbf{Model reasoning}\par\smallskip
\noindent{\raggedright\footnotesize\ttfamily We need to understand the puzzle. Grid is 5x5 rows? Let's map coordinates. Provided grid rows list:\par Row0: ['+', '+', '+', '+', '+']\newline Row1: ['V', '*-Y', '+', '*-Y', '+']\newline Row2: ['V', '+', '+', '+', '+']\newline Row3: ['V', '*-K', '+', '*-K', '+']\newline Row4: ['V', 'L', 'E', '+', '+']\par Current position [1 4] presumably (x, y). Need to confirm coordinate system: likely (x column, y row) zero-index? Starting with positions like row indices. The current position is at row4, column1? Actually [1 4] meaning x=1, y=4. Row index 4 is bottom row (row4). Column index 1 within that row is element 'L' (makes sense). So L is at row4 col1.\par Grid includes 'E' at row4 col2 (x=2,y=4). So end node is adjacent to current position horizontally to the right.\par Legal actions list: [0=RIGHT] only move possible (maybe no up etc). So we must move right to reach E. The path will go from L at (1,4) to E at (2,4). That would end puzzle.\par We must output Final: 0 (RIGHT).\par We should ensure that moving right is allowed (not a gap or visited). Row4 col2 currently 'E' which is end; moving into E is allowed, I think. So final step.\par Thus answer: Final: 0\par}
\par\bigskip
\noindent\rule{\linewidth}{0.4pt}\par\medskip
\newpage
\noindent\textbf{Final frame} \quad {\small\textit{(after last move: \textbf{Right})}}
\par\medskip
\begin{center}
\includegraphics[width=0.34\textwidth]{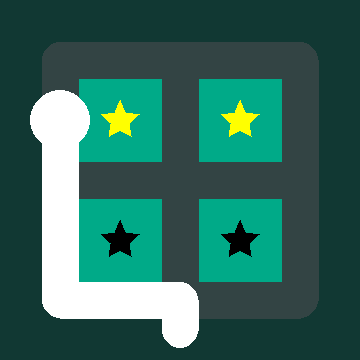}
\\[0.5em]
{\small\itshape terminal state}
\end{center}
\normalsize

\section{Additional Results}

\subsection{Per-Model Rule-Type Analysis}
\label{app:rule_per_model}

\begin{wrapfigure}{R}{0.5\textwidth}
\centering
\includegraphics[width=0.8\linewidth]{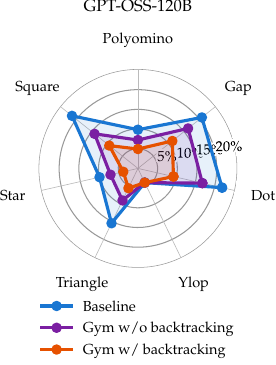}
    \caption{Rule-specific accuracy (\%) across all seven rule types for GPT-OSS 120B under each of the three evaluation settings: SPaRC, Spatial-Gym, and Gym with backtracking.}
\label{fig:radar_gptoss}
\end{wrapfigure}

\Cref{fig:radar_gptoss} shows the rule-type profile for GPT-OSS 120B, which exhibited the largest performance variation across evaluation formats. The radar chart illustrates a striking consistency in the model's relative strengths and weaknesses. 
The geometric shape of the performance polygon remains nearly identical across the Baseline, Gym without backtracking, and Gym with backtracking settings, merely scaling inward as overall accuracy degrades. 

Specifically, the model consistently performs best on the Gap, Square, and Dot constraints. In the Baseline setting, accuracy on these rules peaks at or slightly above $20\%$, but uniformly contracts to roughly $10\%$ to $15\%$ in the standard Gym setting, and drops even further inward when backtracking is enabled. Conversely, the Ylop constraint represents a rigid lower bound; performance on this complex shape-subtraction rule remains exceptionally low (hovering near or below the $5\%$ mark) under all three formats, showing almost no variation. Mid-tier rules, such as Triangles, Stars, and Polyominoes, follow the exact same proportional degradation trend. 

This uniform contraction indicates that the accuracy differences GPT-OSS 120B experiences across settings reflect a global shift in its ability to navigate the sequential format, rather than a selective improvement or disadvantage regarding any specific spatial constraint.
\newpage
\subsection{Backtracking Details}
\label{app:backtracking_details}

\begin{wrapfigure}[21]{r}{0.5\textwidth}
\centering
\includegraphics[width=0.48\textwidth]{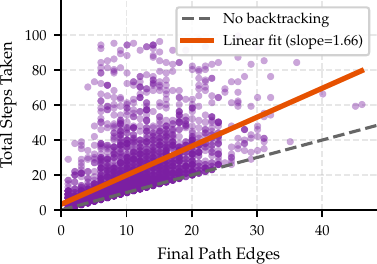}
    \caption{Total steps taken versus final path length (in edges) for all puzzle attempts in the Gym with backtracking setting. The dashed line denotes the no-backtracking baseline (slope\,=\,1); the orange line shows the linear fit (slope\,=\,1.66).}
\label{fig:backtracking_scatter}
\end{wrapfigure}

\Cref{fig:backtracking_scatter} visualizes the relationship between the total number of environment steps taken and the final length of the generated path (in edges) for all puzzle attempts within the Gym with backtracking. The dashed gray line establishes the no-backtracking baseline (slope\,=\,1), representing a direct trajectory where every action consecutively extends the path. The orange line, depicting a linear fit with a slope of 1.66, quantifies the average utilization of the backtrack mechanism. This slope indicates that models generally dedicate approximately two-thirds of their actions to forward progress and one-third to retracing their steps. 

Beyond the average, the plotted distribution provides deeper insight into how models behave when given the ability to revise their decisions. A dense cluster of points lies immediately along the baseline, indicating that models frequently commit to their initial decisions and advance without utilizing the correction mechanism at all. In stark contrast, the wide vertical dispersion above the diagonal highlights episodes of extensive backtracking. In these instances, models actively leverage the backtracking feature to undo sequences of moves, occasionally reaching the environment's 100-step maximum. However, these high step counts generally correspond to short final paths. This behavioral pattern suggests that instead of using the backtrack ability to explore alternative, longer routes necessary to navigate complex rule constraints, models tend to use it to recursively prune their paths, returning to earlier states without successfully advancing deeper into the puzzle grid.

\subsection{Vision Ablation Details}
\label{app:vision_details}

\begin{wrapfigure}{R}{0.5\textwidth}
\centering
\includegraphics[width=0.48\textwidth]{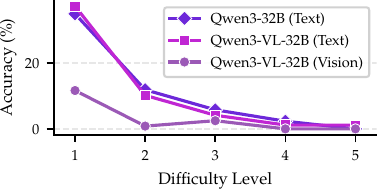}
    \caption{Accuracy (\%) broken down by difficulty level (1--5) for Qwen3-32B (text), Qwen3-VL-32B (text), and Qwen3-VL-32B (vision).}
\label{fig:vision_difficulty}
\end{wrapfigure}

\Cref{fig:vision_difficulty} provides a breakdown of the vision ablation experiments across the five puzzle difficulty levels. As expected, all three configurations show a decline in accuracy as puzzle complexity increases, ultimately converging to near-zero performance at levels 4 and 5. This convergence indicates that the most complex spatial constraints exceed current model capabilities regardless of the provided input modality.

Notably, the performance of the two text-based evaluations (the standard Qwen3-32B model and the vision-language variant Qwen3-VL-32B) tracks almost identically across the entire difficulty spectrum. Both models achieve their peak accuracy of approximately 35\% on the easiest puzzles and decline in tandem. This confirms that integrating visual processing architectures into the VL model does not degrade its underlying text-based spatial reasoning capacity.

In contrast, supplying rendered puzzle images to the Qwen3-VL-32B model severely impairs its performance right from the start. Even at difficulty level 1, the vision-enabled agent achieves only about 12\% accuracy, roughly a two-thirds reduction compared to its text-only counterpart. By difficulty level 2, the vision model's performance collapses to near zero, whereas the text-based models still maintain a solve rate above 10\%. This massive divergence on the easiest puzzles underscores that current multimodal models struggle to reliably extract structured, relational logic from raw pixels, becoming overwhelmed by visual input on tasks they can otherwise partially solve when the state is represented purely as text.

\subsection{Reasoning Mode}
\label{app:reasoning_mode}

Reasoning models with extended CoT outperformed instruction-tuned models in the baseline evaluation \citep{kaesberg2025sparcspatialpathfindingreasoning}.
In \sparcgym{}, direction choice is limited to one of the four possible ones. %
We test whether extended reasoning remains important at this low per-step complexity, or whether the structured environment compensates for its absence.
We compare Qwen~3 14B and 32B in reasoning mode (extended thinking enabled) versus non-reasoning mode (standard output) across baseline and \sparcgym{}.

\begin{figure}[h]
    \includegraphics[width=\textwidth]{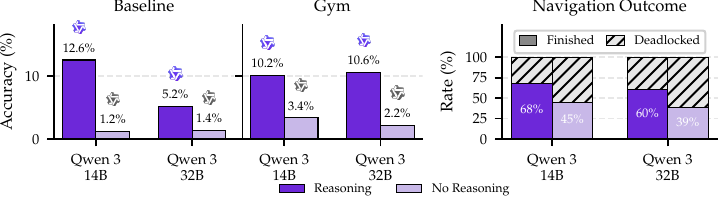}
    \vspace{-18pt}
    \caption{Reasoning versus non-reasoning mode ablation for Qwen~3 14B and Qwen~3 32B. Accuracy (\%) is shown for both baseline and \sparcgym{} settings (left and center), and completion rate (\%) is shown for the Gym setting (right). Each group compares the reasoning-enabled (dark) and reasoning-disabled (light) variants of the same model.}
\label{fig:reasoning_mode}
\end{figure}

\Cref{fig:reasoning_mode} shows accuracy and completion for both modes across the two settings.
In the baseline evaluation, reasoning mode yields large gains.
Qwen~3 14B achieves 12.6\% accuracy with reasoning enabled, but only 1.2\% without, and Qwen~3 32B reaches 5.2\% but only 1.4\% without.
In \sparcgym{} we still see the same difference, but not as prominent.
The 14B model scores 10.2\% with reasoning versus 3.4\% without, and the 32B model achieves 10.6\% versus 2.2\%.
Reasoning mode provides a $3-5 \times$ advantage even in the step-by-step setting.
Completion follows a similar pattern, as reasoning-enabled models finish 60--68\% of puzzles while those without reasoning reach only 39--45\%.
Non-reasoning models become deadlocked more often, suggesting they cannot reliably plan even a single step without extended thinking.

The structured Gym environment does not compensate for the absence of reasoning.
\sparcgym{} partially closes the gap between reasoning and non-reasoning variants relative to the baseline, but a $3$--$5\times$ difference persists.
Non-reasoning models fail at both rule satisfaction and basic navigation, confirming that spatial constraint reasoning requires deliberate multi-step reasoning that the structure of the sequential environment alone cannot replace.
This aligns with \citet{sprague2025cotcotchainofthoughthelps}, who show that CoT provides the largest gains on tasks requiring sequential constraint evaluation rather than single-step pattern matching.

\subsection{Inter-Model Agreement}
\label{app:model_agreement}

\begin{figure}[h]
\centering
\includegraphics[width=\linewidth]{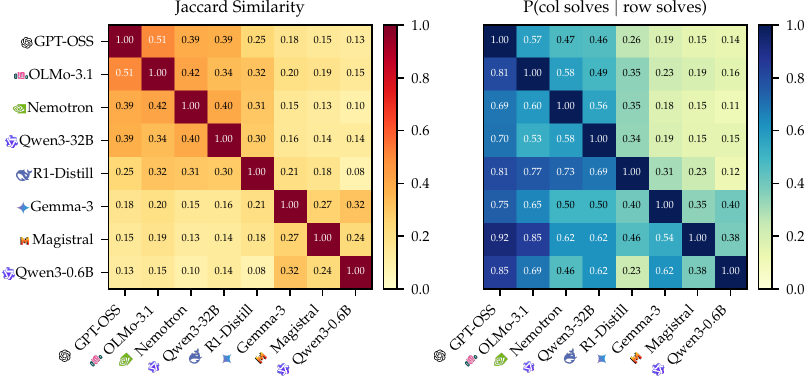}
    \caption{Inter-model agreement matrices for the Spatial-Gym setting. (left) Jaccard similarity between each pair of models' solved puzzle sets. (right) Conditional probability $P(\text{col solves} \mid \text{row solves})$, where entry $(i,j)$ denotes the fraction of puzzles solved by row model $i$ that are also solved by column model $j$.}
\label{fig:agreement}
\end{figure}

\Cref{fig:agreement} visualizes the agreement between the different models in the Spatial-Gym evaluation format. The Jaccard similarity matrix on the left measures the direct overlap between the sets of puzzles successfully solved by each model pair. On the right, the conditional probability matrix illustrates the likelihood that a particular model successfully solves a puzzle given that another specific model has already solved it. The high conditional probabilities observed in the rows of the strongest models, such as GPT-OSS 120B and OLMo 3.1 32B, suggest that they generally solve a superset of the puzzles completed by weaker models. This pattern implies shared failure modes and similar reasoning limitations across different architectures when navigating spatial constraints.

\subsection{Unique Solves Analysis}
\label{app:unique_solves}

\begin{wrapfigure}{R}{0.5\textwidth}
\centering
\includegraphics[width=0.48\textwidth]{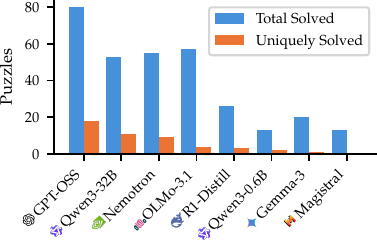}
    \caption{Total number of puzzles solved (blue) versus puzzles solved exclusively by that model and no other (orange) in the Spatial-Gym setting, for each of the eight evaluated models.}
\label{fig:unique_solves}
\end{wrapfigure}

\Cref{fig:unique_solves} shows a comparison of the total number of puzzles solved by each model (blue bars) with the subset of puzzles uniquely solved by that model alone (orange bars) within the Spatial-Gym setting. Unsurprisingly, the top-performing model, GPT-OSS 120B, dominates both metrics, solving exactly 80 puzzles in total and accounting for the largest share of uniquely solved instances (nearly 20). 

However, a closer examination of the mid-tier reasoning models reveals a non-linear relationship between overall spatial capability and solution uniqueness. For instance, while OLMo 3.1 32B solves slightly more puzzles in total than both Nemotron 49B and Qwen3-32B, it produces fewer than half as many unique solves. This shows that OLMo mostly solves the same puzzles as the other top models. In contrast, Qwen3-32B and Nemotron use unique approaches that allow them to solve specific puzzles that the others cannot.

Furthermore, it is notable that almost all of the weaker models (including the small Qwen3-0.6B and the non-reasoning Gemma 3 27B) manage to uniquely solve at least one or two specific puzzles that the stronger models fail on entirely. The only exception is Magistral, whose successful trajectories are entirely subsumed by other agents. This distribution suggests that while overall spatial capability tightly correlates with model strength, differing architectural biases or random walk variations during step-by-step generation occasionally allow weaker agents to find valid paths that highly confident models systematically overlook. Consequently, this diversity implies that multi-agent ensemble methods \citep{becker-etal-2025-mallm} or majority-voting strategies \citep{kaesberg-etal-2025-voting} could yield measurable performance gains in constraint-heavy spatial environments.

\subsection{Performance Scaling Across Difficulty}
\label{app:accuracy_difficulty}

\begin{figure}[h]
    \centering
    \includegraphics[width=\linewidth]{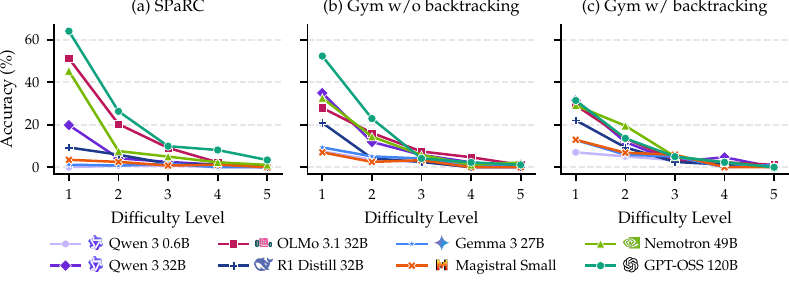}
    \caption{Accuracy (\%) by difficulty level (1--5) for all eight models across the three evaluation settings: (a) original SPaRC, (b) Gym without backtracking, and (c) Gym with backtracking. Each line corresponds to one model.}
    \label{fig:accuracy_by_difficulty}
\end{figure}

\Cref{fig:accuracy_by_difficulty} demonstrates how model performance degrades as puzzle difficulty increases from level 1 to level 5 across the three distinct evaluation settings. Across all formats, accuracy drops substantially on harder puzzle configurations. In the baseline SPaRC setting, certain models exhibit sharp, non-monotonic drops, which heavily conflate formatting difficulties with reasoning failures. In contrast, the two Spatial-Gym settings yield notably smoother scaling curves, confirming that the stepwise environment successfully isolates spatial reasoning capabilities by eliminating strict formatting requirements. At the highest difficulty level, accuracy universally converges near zero, emphasizing a rigid capability limit that current models cannot bypass regardless of the prompting or interaction format.

\subsection{Accuracy versus Reasoning Effort}
\label{app:accuracy_tokens}

\begin{figure}[h]
    \centering
    \includegraphics[width=0.8\linewidth]{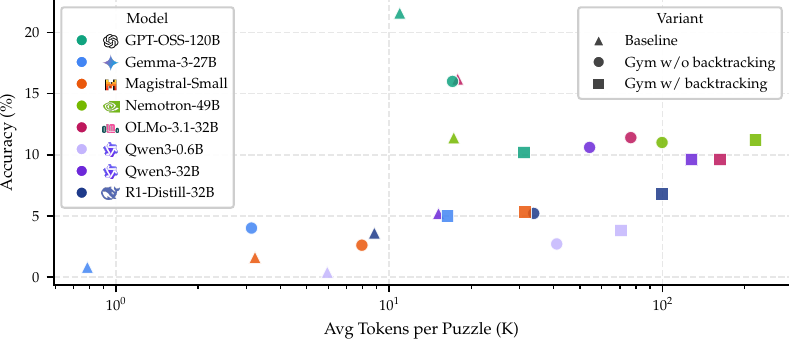}
    \caption{Accuracy (\%) versus average token count per puzzle (K, log scale) for all eight models across the three evaluation settings (Baseline: $\blacktriangle$, Spatial-Gym: \raisebox{-0.8ex}{\scalebox{2}{\textbullet}}, Backtracking: $\blacksquare$). Each color denotes a distinct model.}
    \label{fig:accuracy_vs_tokens}
\end{figure}

\Cref{fig:accuracy_vs_tokens} maps overall model accuracy against the average token count generated per puzzle in a logarithmic scale. The sequential Spatial-Gym formats inherently require substantially more token generation than the one-shot SPaRC baseline due to the continuous back-and-forth interactions. However, producing an increased volume of tokens does not reliably translate into higher reasoning accuracy. This disconnect becomes especially apparent in the backtracking setting, where models generate the most tokens. Instead of leveraging this extra computational effort to explore complex constraint satisfaction, models tend to spend their tokens navigating toward shorter, simpler paths, which ultimately harms the performance of stronger agents.

\subsection{Token Usage}
\label{app:token_usage}

\begin{figure}[h]
    \centering
    \includegraphics[width=\linewidth]{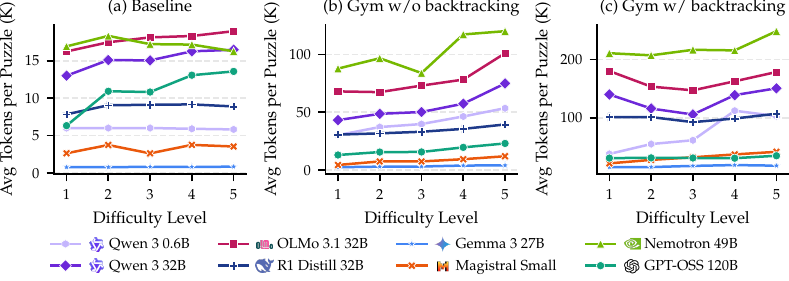}
    \vspace{-18pt}
    \caption{Average token count per puzzle (in thousands) across difficulty levels 1--5 for all eight models under (a) baseline, (b) Gym without backtracking, and (c) Gym with backtracking. Each line corresponds to one model.}
    \label{fig:token_count_difficulty}
\end{figure}

\Cref{fig:token_count_difficulty} shows the average token count per puzzle across difficulty levels (note that the difficulty level is the ceiling of the continuous difficulty score).
(a) The baseline shows no consistent scaling trend.
(b) Under \sparcgym{}, all models show a clearer upward trend, with token consumption increasing more than $5\times$ compared to baseline.
(c) With backtracking, token counts roughly double, though the scaling with difficulty flattens.

\section{Training of Reasoning Agents}
\label{app:rl_training}

Since \sparcgym{} is implemented as a Gymnasium-compatible sequential decision environment, it can also serve as a testbed for reinforcement learning. While the main paper focuses on evaluation, we additionally conduct preliminary RL experiments with Qwen2.5-3B-Instruct~\citep{qwen2.5} to investigate whether task-specific interaction and reward optimization can improve spatial reasoning performance. We choose this model scale due to limited compute resources and the high cost of RL training in \sparcgym{}, where each rollout involves relatively long textual state inputs and long-horizon interactions. Our implementation is based on RAGEN~\citep{ragen}, and we use the default \sparcgym{} setting with backtracking disabled.

At each environment step, the model receives a system prompt consisting of the general task instructions (\Cref{app:prompt-gym}) together with the current textual state returned by the environment. This state includes the current grid configuration, the traced path so far, the agent's current position, and the set of legal moves. Conditioned on this input, the model generates a structured response in a ReAct-style format~\citep{yao2022react}, where the intermediate reasoning is enclosed in \texttt{<think> </think>} tags and the selected action is enclosed in \texttt{<answer> </answer>} tags. The action extracted from the \texttt{<answer>} block is then parsed and executed in the environment as the agent's next move.

Because interaction with the environment depends on successful parsing, only responses that follow the prescribed format can be executed correctly. To encourage stable interaction, we introduce an additional penalty for invalid outputs, such as missing tags or an unparsable action. This penalty is used only during training and is intended to discourage malformed responses rather than poor decision-making itself. This design allows the model to produce free-form intermediate reasoning while still interfacing with the environment through a discrete executable action. As the environment state is updated after each step, trajectories are generated auto-regressively through repeated model-environment interaction.

We train directly from the base model, without supervised fine-tuning on expert trajectories, using Proximal Policy Optimization (PPO)~\citep{schulman2017proximal} with Generalized Advantage Estimation (GAE). Specifically, trajectories are sampled on-policy through interaction with the environment. For each trajectory, the process reward, outcome reward, and format penalty are summed to form the training reward signal. PPO updates are then applied to increase the likelihood of high-return responses while constraining policy drift between updates. We set the maximum number of interaction steps to 31. For these preliminary experiments, we train the model only on difficulty-1 and difficulty-2 puzzles for 200 training steps, since easier instances provide a more stable starting point for RL from a base model by increasing the frequency of successful trajectories and making credit assignment more tractable.
The overall results are shown in \Cref{tab:rl_results}.

\begin{table*}[h]
\centering
\small
\setlength{\tabcolsep}{4pt}
\begin{tabular}{lcccccccccc}
\toprule
\textbf{Model} & \textbf{Acc.} & \textbf{\#Avg. Steps} & \textbf{\#Min. Steps} & \textbf{\#Max. Steps} & \textbf{Term.}  & \textbf{D1}  & \textbf{D2}  & \textbf{D3}  & \textbf{D4} & \textbf{D5}  \\
\midrule
Base    & 2.6 & 21.3 & 1 & 75 & 35.0 & 8.1 & 4.2 & 0.8 & 0.0 & 0.0 \\
Trained & 3.6 & 21.2 & 1 & 86 & 33.6 & 9.3 & 5.1 & 2.5 & 1.2 & 0.0 \\
\bottomrule
\end{tabular}
\caption{Performance of the base model and the RL-trained model on \sparcgym{}. \textbf{Acc.} denotes the percentage of puzzles solved successfully. \textbf{Avg. Steps} is the average number of interaction steps over all puzzles in the test set, while \textbf{Min. Steps} and \textbf{Max. Steps} denote the minimum and maximum numbers of steps, respectively. \textbf{Term.} denotes the percentage of trajectories that reach the target endpoint, regardless of whether all constraints are satisfied. \textbf{D1}--\textbf{D5} report accuracy for puzzle difficulty levels 1--5.}
\label{tab:rl_results}
\end{table*}

The overall results are shown in \Cref{tab:rl_results}. Reinforcement learning yields a modest but consistent improvement over the base model. Overall accuracy increases from 2.6\% to 3.6\%, showing that direct policy optimization in \sparcgym{} can improve performance even when training starts from the base model without supervised trajectory imitation. The improvement is also reflected in the difficulty-wise breakdown: the trained model performs better on difficulty levels 1 through 4, with gains from 8.1\% to 9.3\% on difficulty-1 puzzles, from 4.2\% to 5.1\% on difficulty-2 puzzles, from 0.8\% to 2.5\% on difficulty-3 puzzles, and from 0.0\% to 1.2\% on difficulty-4 puzzles, while both models fail on difficulty-5 puzzles.

At the same time, the gains remain small in absolute terms. The average number of interaction steps is nearly unchanged (21.3 vs.\ 21.2), and the termination rate slightly decreases from 35.0\% to 33.6\%, suggesting that the trained model is not simply learning to reach the endpoint more often, but only modestly improving its ability to satisfy the full set of puzzle constraints. More broadly, the persistently low performance, even after task-specific RL training, highlights the difficulty of the environment and further supports the value of \sparcgym{} as a benchmark for spatial reasoning. These results suggest that the challenge posed by \sparcgym{} is not merely due to prompt format or lack of interaction, but reflects a more fundamental gap in current models' spatial reasoning and planning abilities. Overall, our preliminary RL results show that \sparcgym{} is not only a viable training environment but also a demanding benchmark whose difficulty cannot be overcome by straightforward policy optimization alone.

\section{Full Tabular Results}
\label{app:full_tabular_results}

This appendix provides the comprehensive numerical data for all experiments discussed in the main text. The tables detail both the overall performance metrics across difficulty levels and the granular performance across specific rule types for the eight evaluated models. \Cref{tab:sparc-results} presents the baseline performance on the original SPaRC one-shot evaluation, including overall accuracy, solve rates for difficulty levels 1 through 5, and the average number of steps taken per puzzle. \Cref{tab:sparc-gym-results} details the exact metrics for the standard Spatial-Gym setting without backtracking, again providing the solve rates for the five difficulty levels alongside the average step counts. \Cref{tab:sparc-gym-tb-results} displays the corresponding results for the Spatial-Gym setting with the backtracking feature enabled, allowing for a direct assessment of how backtracking influences step counts and accuracy across the difficulty spectrum.

\begin{table*}[h]
  \centering
  \small
  \begin{tabular}{cl | c | ccccc | c}
    \toprule
    & \textbf{Model} & \textbf{Accuracy} & \textbf{D1} & \textbf{D2} & \textbf{D3} & \textbf{D4} & \textbf{D5} & \textbf{Avg.\ Steps} \\
    \midrule
    \raisebox{-0.2\height}{\includegraphics[width=1em]{images/logos/openai.png}} & GPT-OSS 120B & \textbf{21.6} & \textbf{64.0} & \textbf{26.3} & \textbf{9.9} & \textbf{8.1} & \textbf{3.4} & 19.8 \\
    \raisebox{-0.2\height}{\includegraphics[width=1em]{images/logos/nvidia.png}} & Nemotron 49B & 11.4 & 45.3 & 7.6 & 5.0 & 2.3 & 1.1 & 20.0 \\
    \raisebox{-0.2\height}{\includegraphics[width=1em]{images/logos/qwen.png}} & Qwen 3 32B & 5.2 & 19.8 & 4.2 & 2.5 & 1.2 & 0.0 & 20.0 \\
    \raisebox{-0.2\height}{\includegraphics[width=1em]{images/logos/olmo.png}} & OLMo 3.1 32B & 16.2 & 51.2 & 20.3 & 9.1 & 2.3 & 0.0 & 16.1 \\
    \raisebox{-0.2\height}{\includegraphics[width=1em]{images/logos/deepseek.png}} & R1 Distill 32B & 3.6 & 9.3 & 5.9 & 1.7 & 1.2 & 0.0 & 19.3 \\
    \raisebox{-0.2\height}{\includegraphics[width=1em]{images/logos/gemma.png}} & Gemma 3 27B & 0.8 & 1.2 & 0.8 & 1.7 & 0.0 & 0.0 & 17.6 \\
    \raisebox{-0.2\height}{\includegraphics[width=1em]{images/logos/mistral.png}} & Magistral Small & 1.6 & 3.5 & 2.5 & 0.8 & 1.2 & 0.0 & 20.3 \\
    \raisebox{-0.2\height}{\includegraphics[width=1em]{images/logos/qwen.png}} & Qwen 3 0.6B & 0.4 & 0.0 & 0.8 & 0.8 & 0.0 & 0.0 & 15.5 \\
    \bottomrule
  \end{tabular}
    \caption{SPaRC results. Overall accuracy (\%) and per-difficulty solve rate (\%, D1--D5) for all eight models, along with the average number of steps taken per puzzle. Best results in \textbf{bold}.}
    \label{tab:sparc-results}
\end{table*}

\begin{table*}[h]
  \centering
  \small
  \begin{tabular}{cl | c | ccccc | c}
    \toprule
    & \textbf{Model} & \textbf{Accuracy} & \textbf{D1} & \textbf{D2} & \textbf{D3} & \textbf{D4} & \textbf{D5} & \textbf{Avg.\ Steps} \\
    \midrule
    \raisebox{-0.2\height}{\includegraphics[width=1em]{images/logos/openai.png}} & GPT-OSS 120B & \textbf{16.0} & \textbf{52.3} & \textbf{22.9} & 4.1 & 2.3 & 1.1 & 14.7 \\
    \raisebox{-0.2\height}{\includegraphics[width=1em]{images/logos/nvidia.png}} & Nemotron 49B & 11.0 & 32.6 & 14.4 & 5.8 & 1.2 & \textbf{2.2} & 18.1 \\
    \raisebox{-0.2\height}{\includegraphics[width=1em]{images/logos/qwen.png}} & Qwen 3 32B & 10.6 & 34.9 & 11.9 & 5.8 & 2.3 & 0.0 & 15.7 \\
    \raisebox{-0.2\height}{\includegraphics[width=1em]{images/logos/olmo.png}} & OLMo 3.1 32B & 11.4 & 27.9 & 16.1 & \textbf{7.4} & \textbf{4.7} & 1.1 & 14.8 \\
    \raisebox{-0.2\height}{\includegraphics[width=1em]{images/logos/deepseek.png}} & R1 Distill 32B & 5.2 & 20.9 & 4.2 & 2.5 & 0.0 & 0.0 & 13.6 \\
    \raisebox{-0.2\height}{\includegraphics[width=1em]{images/logos/gemma.png}} & Gemma 3 27B & 4.0 & 9.3 & 5.1 & 4.1 & 1.2 & 0.0 & 11.8 \\
    \raisebox{-0.2\height}{\includegraphics[width=1em]{images/logos/mistral.png}} & Magistral Small & 2.6 & 7.0 & 2.5 & 3.3 & 0.0 & 0.0 & 12.8 \\
    \raisebox{-0.2\height}{\includegraphics[width=1em]{images/logos/qwen.png}} & Qwen 3 0.6B & 2.7 & 7.4 & 3.5 & 2.5 & 0.0 & 0.0 & 14.4 \\
    \bottomrule
  \end{tabular}
    \caption{\sparcgym{} results. Overall accuracy (\%) and per-difficulty solve rate (\%, D1--D5) for all eight models, along with the average number of steps taken per puzzle. Best results in \textbf{bold}.}
  \label{tab:sparc-gym-results}
\end{table*}

\begin{table*}[h]
  \centering
  \small
  \begin{tabular}{cl | c | ccccc | c}
    \toprule
    & \textbf{Model} & \textbf{Accuracy} & \textbf{D1} & \textbf{D2} & \textbf{D3} & \textbf{D4} & \textbf{D5} & \textbf{Avg.\ Steps} \\
    \midrule
    \raisebox{-0.2\height}{\includegraphics[width=1em]{images/logos/openai.png}} & GPT-OSS 120B & 10.2 & \textbf{31.4} & 13.6 & 5.0 & 2.3 & 0.0 & 20.3 \\
    \raisebox{-0.2\height}{\includegraphics[width=1em]{images/logos/nvidia.png}} & Nemotron 49B & \textbf{11.2} & 29.1 & \textbf{19.5} & 5.0 & 2.3 & 0.0 & 33.4 \\
    \raisebox{-0.2\height}{\includegraphics[width=1em]{images/logos/qwen.png}} & Qwen 3 32B & 9.6 & \textbf{31.4} & 11.9 & 2.5 & \textbf{4.7} & 0.0 & 29.3 \\
    \raisebox{-0.2\height}{\includegraphics[width=1em]{images/logos/olmo.png}} & OLMo 3.1 32B & 9.6 & 29.1 & 13.6 & 3.3 & 2.3 & \textbf{1.1} & 24.6 \\
    \raisebox{-0.2\height}{\includegraphics[width=1em]{images/logos/deepseek.png}} & R1 Distill 32B & 6.8 & 22.1 & 9.3 & 2.5 & 1.2 & 0.0 & 34.4 \\
    \raisebox{-0.2\height}{\includegraphics[width=1em]{images/logos/gemma.png}} & Gemma 3 27B & 5.0 & 12.8 & 5.9 & 5.0 & 0.0 & \textbf{1.1} & 57.4 \\
    \raisebox{-0.2\height}{\includegraphics[width=1em]{images/logos/mistral.png}} & Magistral Small & 5.3 & 12.9 & 6.8 & \textbf{5.9} & 0.0 & 0.0 & 45.4 \\
    \raisebox{-0.2\height}{\includegraphics[width=1em]{images/logos/qwen.png}} & Qwen 3 0.6B & 3.8 & 7.0 & 5.1 & 3.3 & 2.3 & \textbf{1.1} & 20.2 \\
    \bottomrule
  \end{tabular}
    \caption{\sparcgym{} with backtracking results. Overall accuracy (\%) and per-difficulty solve rate (\%, D1--D5) for all eight models, along with the average number of steps taken per puzzle. Best results in \textbf{bold}.}
  \label{tab:sparc-gym-tb-results}
\end{table*}

\clearpage
Furthermore, the results are broken down by the specific spatial constraints present in the puzzles, which include Dot, Gap, Polyomino, Ylop, Star, Square, and Triangle. \Cref{tab:sparc-rule-rates} lists the rule-specific accuracies for the SPaRC baseline setting. \Cref{tab:sparc-gym-rule-rates} isolates the rule-specific performance within the standard Spatial-Gym environment. Finally, \Cref{tab:sparc-gym-tb-rule-rates} shows the rule-specific accuracy when backtracking is enabled in the Gym environment.

\begin{table*}[h]
  \centering
  \small
  \begin{tabular}{cl |  c c c c c c c}
    \toprule
    & \textbf{Model} & \textbf{Dot} & \textbf{Gap} & \textbf{Polyomino} & \textbf{Ylop} & \textbf{Star} & \textbf{Square} & \textbf{Triangle} \\
    \midrule
    \raisebox{-0.2\height}{\includegraphics[width=1em]{images/logos/openai.png}} & GPT-OSS 120B & \textbf{21.9} & \textbf{20.8} & \textbf{9.8} & \textbf{4.0} & \textbf{10.0} & \textbf{21.4} & \textbf{15.5} \\
    \raisebox{-0.2\height}{\includegraphics[width=1em]{images/logos/nvidia.png}} & Nemotron 49B & 13.7 & 9.6 & 5.2 & \textbf{4.0} & 2.9 & 11.0 & 5.2 \\
    \raisebox{-0.2\height}{\includegraphics[width=1em]{images/logos/qwen.png}} & Qwen 3 32B & 5.5 & 5.1 & 2.3 & 0.0 & 1.4 & 4.2 & 2.1 \\
    \raisebox{-0.2\height}{\includegraphics[width=1em]{images/logos/olmo.png}} & OLMo 3.1 32B & 17.5 & 15.0 & 6.6 & 0.0 & 4.3 & 16.3 & 10.7 \\
    \raisebox{-0.2\height}{\includegraphics[width=1em]{images/logos/deepseek.png}} & R1 Distill 32B & 2.4 & 3.8 & 0.3 & 0.0 & 2.4 & 3.4 & 2.1 \\
    \raisebox{-0.2\height}{\includegraphics[width=1em]{images/logos/gemma.png}} & Gemma 3 27B & 0.3 & 0.3 & 1.0 & 0.0 & 0.0 & 1.1 & 0.0 \\
    \raisebox{-0.2\height}{\includegraphics[width=1em]{images/logos/mistral.png}} & Magistral Small & 0.3 & 1.6 & 1.6 & 0.0 & 1.0 & 1.7 & 0.4 \\
    \raisebox{-0.2\height}{\includegraphics[width=1em]{images/logos/qwen.png}} & Qwen 3 0.6B & 0.0 & 0.3 & 0.7 & 0.0 & 0.0 & 0.6 & 0.0 \\
    \bottomrule
  \end{tabular}
    \caption{SPaRC accuracy (\%) broken down by rule type for all eight models. Best results in \textbf{bold}.}
  \label{tab:sparc-rule-rates}
\end{table*}

\begin{table*}[h]
  \centering
  \small
  \begin{tabular}{cl |  c c c c c c c}
    \toprule
    & \textbf{Model} & \textbf{Dot} & \textbf{Gap} & \textbf{Polyomino} & \textbf{Ylop} & \textbf{Star} & \textbf{Square} & \textbf{Triangle} \\
    \midrule
    \raisebox{-0.2\height}{\includegraphics[width=1em]{images/logos/openai.png}} & GPT-OSS 120B & \textbf{16.8} & \textbf{16.3} & \textbf{7.2} & \textbf{4.0} & \textbf{7.1} & \textbf{14.1} & \textbf{9.0} \\
    \raisebox{-0.2\height}{\includegraphics[width=1em]{images/logos/nvidia.png}} & Nemotron 49B & 9.9 & 10.9 & 3.0 & 0.0 & 6.2 & 10.7 & \textbf{9.0} \\
    \raisebox{-0.2\height}{\includegraphics[width=1em]{images/logos/qwen.png}} & Qwen 3 32B & 9.2 & 9.6 & 5.6 & \textbf{4.0} & 4.8 & 9.6 & 3.9 \\
    \raisebox{-0.2\height}{\includegraphics[width=1em]{images/logos/olmo.png}} & OLMo 3.1 32B & 9.6 & 11.2 & 5.2 & \textbf{4.0} & 6.7 & 11.8 & 7.3 \\
    \raisebox{-0.2\height}{\includegraphics[width=1em]{images/logos/deepseek.png}} & R1 Distill 32B & 4.8 & 5.8 & 1.0 & 0.0 & 1.4 & 4.8 & 2.1 \\
    \raisebox{-0.2\height}{\includegraphics[width=1em]{images/logos/gemma.png}} & Gemma 3 27B & 1.4 & 3.8 & 3.0 & \textbf{4.0} & 2.4 & 3.4 & 0.4 \\
    \raisebox{-0.2\height}{\includegraphics[width=1em]{images/logos/mistral.png}} & Magistral Small & 0.7 & 2.6 & 1.3 & 0.0 & 1.4 & 2.3 & 0.4 \\
    \raisebox{-0.2\height}{\includegraphics[width=1em]{images/logos/qwen.png}} & Qwen 3 0.6B & 0.4 & 2.3 & 2.7 & 0.0 & 1.5 & 2.1 & 0.4 \\
    \bottomrule
  \end{tabular}
    \caption{\sparcgym{} accuracy (\%) broken down by rule type for all eight models. Best results in \textbf{bold}.}
  \label{tab:sparc-gym-rule-rates}
\end{table*}

\begin{table*}[h]
  \centering
  \small
  \begin{tabular}{cl |  c c c c c c c}
    \toprule
    & \textbf{Model} & \textbf{Dot} & \textbf{Gap} & \textbf{Polyomino} & \textbf{Ylop} & \textbf{Star} & \textbf{Square} & \textbf{Triangle} \\
    \midrule
    \raisebox{-0.2\height}{\includegraphics[width=1em]{images/logos/openai.png}} & GPT-OSS 120B & 9.2 & 11.2 & 4.9 & 4.0 & 3.8 & 9.3 & 5.6 \\
    \raisebox{-0.2\height}{\includegraphics[width=1em]{images/logos/nvidia.png}} & Nemotron 49B & \textbf{10.3} & \textbf{11.8} & 3.0 & 4.0 & 5.7 & \textbf{10.4} & \textbf{9.0} \\
    \raisebox{-0.2\height}{\includegraphics[width=1em]{images/logos/qwen.png}} & Qwen 3 32B & 8.2 & 9.9 & 3.9 & 4.0 & 4.3 & 8.7 & 5.2 \\
    \raisebox{-0.2\height}{\includegraphics[width=1em]{images/logos/olmo.png}} & OLMo 3.1 32B & 7.5 & 9.6 & 3.6 & 0.0 & 4.8 & 9.0 & 6.0 \\
    \raisebox{-0.2\height}{\includegraphics[width=1em]{images/logos/deepseek.png}} & R1 Distill 32B & 6.2 & 7.3 & 2.3 & 4.0 & 3.3 & 6.5 & 3.0 \\
    \raisebox{-0.2\height}{\includegraphics[width=1em]{images/logos/gemma.png}} & Gemma 3 27B & 2.4 & 4.5 & 3.3 & 0.0 & 3.8 & 5.1 & 1.3 \\
    \raisebox{-0.2\height}{\includegraphics[width=1em]{images/logos/mistral.png}} & Magistral Small & 2.4 & 5.5 & 4.0 & 4.2 & 3.4 & 3.4 & 2.6 \\
    \raisebox{-0.2\height}{\includegraphics[width=1em]{images/logos/qwen.png}} & Qwen 3 0.6B & 2.5 & 8.8 & \textbf{6.0} & \textbf{7.7} & \textbf{7.7} & 8.6 & 1.9 \\
    \bottomrule
  \end{tabular}
    \caption{\sparcgym{} with backtracking accuracy (\%) broken down by rule type for all eight models. Best results in \textbf{bold}.}
  \label{tab:sparc-gym-tb-rule-rates}
\end{table*}

\onecolumn
\section{AI-Usage Card}
\label{app:ai_card}

\makeAIUsageCard

\clearpage
\onecolumn
\hypertarget{annotation}{}
\pagestyle{empty}
\lstset{
  basicstyle=\footnotesize\ttfamily,
  breaklines=true,
  breakatwhitespace=false,
  columns=flexible,
  numbers=none
}

\definecolor{Primary}{RGB}{59, 130, 246}    %
\definecolor{PrimaryDark}{RGB}{30, 64, 175} %
\definecolor{LightBg}{RGB}{239, 246, 255}   %
\definecolor{TextDark}{RGB}{31, 41, 55}     %
\definecolor{TextMuted}{RGB}{107, 114, 128} %

\begin{tikzpicture}[remember picture, overlay]
  \fill[Primary] ([xshift=0cm,yshift=0cm]current page.north west) rectangle ([xshift=\paperwidth,yshift=-0.4cm]current page.north west);
\end{tikzpicture}

\vspace{0.8cm}
\begin{center}
  {\fontsize{22}{26}\selectfont\sffamily\bfseries \textcolor{PrimaryDark}{CiteAssist}}\\[0.2em]
  {\Large\sffamily\scshape \textcolor{TextMuted}{Citation Sheet}}\\[0.8em]
  {\small\sffamily Generated with \href{https://citeassist.uni-goettingen.de/}{\textcolor{Primary}{\texttt{citeassist.uni-goettingen.de}}}
  \CiteAssistCite{}
  }\end{center}

\begin{center}
\vspace{1em}
\begin{tikzpicture}
\draw[Primary, line width=0.6pt] (0,0) -- (\textwidth,0);
\end{tikzpicture}
\vspace{1.2em}
\end{center}

\begin{tcolorbox}[enhanced,
                 frame hidden,
                 boxrule=0pt,
                 borderline west={2pt}{0pt}{Primary},
                 colback=LightBg,
                 sharp corners,
                 breakable,
                 fonttitle=\sffamily\bfseries\large,
                 coltitle=Primary,
                 title=BibTeX Entry,
                 attach title to upper={\vspace{0.2em}\par},
                 left=12pt]
\lstset{
    inputencoding = utf8,  %
    extendedchars = true,  %
    literate      =        %
      {á}{{\'a}}1  {é}{{\'e}}1  {í}{{\'i}}1 {ó}{{\'o}}1  {ú}{{\'u}}1
      {Á}{{\'A}}1  {É}{{\'E}}1  {Í}{{\'I}}1 {Ó}{{\'O}}1  {Ú}{{\'U}}1
      {à}{{\`a}}1  {è}{{\`e}}1  {ì}{{\`i}}1 {ò}{{\`o}}1  {ù}{{\`u}}1
      {À}{{\`A}}1  {È}{{\`E}}1  {Ì}{{\`I}}1 {Ò}{{\`O}}1  {Ù}{{\`U}}1
      {ä}{{\"a}}1  {ë}{{\"e}}1  {ï}{{\"i}}1 {ö}{{\"o}}1  {ü}{{\"u}}1
      {Ä}{{\"A}}1  {Ë}{{\"E}}1  {Ï}{{\"I}}1 {Ö}{{\"O}}1  {Ü}{{\"U}}1
      {â}{{\^a}}1  {ê}{{\^e}}1  {î}{{\^i}}1 {ô}{{\^o}}1  {û}{{\^u}}1
      {Â}{{\^A}}1  {Ê}{{\^E}}1  {Î}{{\^I}}1 {Ô}{{\^O}}1  {Û}{{\^U}}1
      {œ}{{\oe}}1  {Œ}{{\OE}}1  {æ}{{\ae}}1 {Æ}{{\AE}}1  {ß}{{\ss}}1
      {ẞ}{{\SS}}1  {ç}{{\c{c}}}1 {Ç}{{\c{C}}}1 {ø}{{\o}}1  {Ø}{{\O}}1
      {å}{{\aa}}1  {Å}{{\AA}}1  {ã}{{\~a}}1  {õ}{{\~o}}1 {Ã}{{\~A}}1
      {Õ}{{\~O}}1  {ñ}{{\~n}}1  {Ñ}{{\~N}}1  {¿}{{?\`}}1  {¡}{{!\`}}1
      {„}{\quotedblbase}1 {“}{\textquotedblleft}1 {–}{$-$}1
      {°}{{\textdegree}}1 {º}{{\textordmasculine}}1 {ª}{{\textordfeminine}}1
      {£}{{\pounds}}1  {©}{{\copyright}}1  {®}{{\textregistered}}1
      {«}{{\guillemotleft}}1  {»}{{\guillemotright}}1  {Ð}{{\DH}}1  {ð}{{\dh}}1
      {Ý}{{\'Y}}1    {ý}{{\'y}}1    {Þ}{{\TH}}1    {þ}{{\th}}1    {Ă}{{\u{A}}}1
      {ă}{{\u{a}}}1  {Ą}{{\k{A}}}1  {ą}{{\k{a}}}1  {Ć}{{\'C}}1    {ć}{{\'c}}1
      {Č}{{\v{C}}}1  {č}{{\v{c}}}1  {Ď}{{\v{D}}}1  {ď}{{\v{d}}}1  {Đ}{{\DJ}}1
      {đ}{{\dj}}1    {Ė}{{\.{E}}}1  {ė}{{\.{e}}}1  {Ę}{{\k{E}}}1  {ę}{{\k{e}}}1
      {Ě}{{\v{E}}}1  {ě}{{\v{e}}}1  {Ğ}{{\u{G}}}1  {ğ}{{\u{g}}}1  {Ĩ}{{\~I}}1
      {ĩ}{{\~\i}}1   {Į}{{\k{I}}}1  {į}{{\k{i}}}1  {İ}{{\.{I}}}1  {ı}{{\i}}1
      {Ĺ}{{\'L}}1    {ĺ}{{\'l}}1    {Ľ}{{\v{L}}}1  {ľ}{{\v{l}}}1  {Ł}{{\L{}}}1
      {ł}{{\l{}}}1   {Ń}{{\'N}}1    {ń}{{\'n}}1    {Ň}{{\v{N}}}1  {ň}{{\v{n}}}1
      {Ő}{{\H{O}}}1  {ő}{{\H{o}}}1  {Ŕ}{{\'{R}}}1  {ŕ}{{\'{r}}}1  {Ř}{{\v{R}}}1
      {ř}{{\v{r}}}1  {Ś}{{\'S}}1    {ś}{{\'s}}1    {Ş}{{\c{S}}}1  {ş}{{\c{s}}}1
      {Š}{{\v{S}}}1  {š}{{\v{s}}}1  {Ť}{{\v{T}}}1  {ť}{{\v{t}}}1  {Ũ}{{\~U}}1
      {ũ}{{\~u}}1    {Ū}{{\={U}}}1  {ū}{{\={u}}}1  {Ů}{{\r{U}}}1  {ů}{{\r{u}}}1
      {Ű}{{\H{U}}}1  {ű}{{\H{u}}}1  {Ų}{{\k{U}}}1  {ų}{{\k{u}}}1  {Ź}{{\'Z}}1
      {ź}{{\'z}}1    {Ż}{{\.Z}}1    {ż}{{\.z}}1    {Ž}{{\v{Z}}}1  {ž}{{\v{z}}}1
  }
\begin{lstlisting}
@misc{kaesberg2026,
  author={Kaesberg, Lars Benedikt and Yang, Tianyu and Bauer, Niklas and Ruas, Terry and Wahle, Jan Philip and Gipp, Bela},
  title={Mind the Gap Between Spatial Reasoning and Acting! Step-by-Step Evaluation of Agents With Spatial-Gym},
  year={2026},
  month={04},
  topic={nlp}
}
\end{lstlisting}
\end{tcolorbox}

\vfill
\begin{tikzpicture}
\draw[Primary!40, line width=0.4pt] (0,0) -- (\textwidth,0);
\end{tikzpicture}
\begin{center}
\small\sffamily\textcolor{TextMuted}{Generated \today}
\end{center}

\end{document}